\documentclass[twocolumn,prl,tightenlines,superscriptaddress,longbibliography]{revtex4-1}
\usepackage{graphicx}
\usepackage{mlmath}
\usepackage{color}
\makeatletter
\usepackage{dcolumn}% Align table columns on decimal point
\usepackage{bm}% bold math
\usepackage[caption=false]{subfig}
\makeatother

\begin{document}
~

%\title{Modeling Frequency-Principle Phenomenon of Neural Networks}
\title{Linear Frequency Principle Model to Understand the Absence of Overfitting in Neural Networks}

\author{Yaoyu Zhang}
% \affiliation{School of Mathematics, Institute for Advanced Study, Princeton,
% NJ 08540, USA}
\author{Tao Luo}
\author{Zheng Ma}
\author{Zhi-Qin John Xu}
\thanks{xuzhiqin@sjtu.edu.cn, to appear in Chinese Physics Letters.}
\affiliation{School of Mathematical Sciences, Institute of
Natural Sciences, MOE-LSC, and Qing Yuan Research Institute, Shanghai Jiao Tong University, Shanghai, 200240, P.R. China}

\date{\today}
\begin{abstract}
Why heavily parameterized neural networks (NNs) do not overfit the data is an important long standing open question. We propose a phenomenological model of the NN training  to explain this non-overfitting puzzle. Our linear frequency principle (LFP) model accounts for a key dynamical feature of NNs: they learn low frequencies first, irrespective of microscopic details. Theory based on our LFP model shows that  low frequency dominance of target functions is the key condition for the non-overfitting of NNs and is verified by experiments. Furthermore, through an ideal two-layer NN, we unravel how detailed microscopic NN training dynamics statistically gives rise to a LFP model with quantitative prediction power. 

\end{abstract}

\pacs{Valid PACS appear here}

\maketitle

Deep learning, a subfield of machine learning achieving huge success in industrial applications, is experiencing a surge in many areas of science including physics \citep{aurisano_convolutional_2016,zhang_deep_2018,guest_deep_2018,radovic_machine_2018,levine_quantum_2019,carleo_machine_2019,mehta_high-bias_2019}. A typical well-solved problem is supervised learning, where the machine learns a mapping from input $\vx\in \mathbb{R}^d$ to output $\vy\in \mathbb{R}^{d_{\rm o}}$ from a training dataset $S=\{(\vx_i,\vy_i)\}_{i=1}^{n}$. The machine is realized by a deep neural network (DNN) of proper depth $L$ 
\[
f_{\vtheta}(\vx)= \mW^{[L]}\sigma\circ(\cdots\mW^{[2]}\sigma\circ(\mW^{[1]}\vx+\vb^{[1]})+\cdots)+\vb^{[L]},
\]
where $\vtheta=\{\mW^{[l]},\vb^{[l]}\}_{l=1}^L$, $\mW^{[l]}$ are weight matrices, $\vb^{[l]}$ are bias vectors, and $\sigma\circ()$ is an element-wise nonlinear activation function.
%The machine is realized by a deep neural network (DNN) in which the output of the ($l-1$)th layer, denoted by $h_{l-1}(\vx)$, is fed as input into the $l$th layer, i.e., $h_l(\vx)=\sigma\circ(\mW_{l-1}h_{l-1}(\vx)+\vb_{l-1})$, where $\mW_{l-1}$ is a weight matrix, $\vb_{l-1}$ is a bias vector, $h_{0}(\vx)=\vx$ is the input, $\sigma\circ(\cdot)$ is a component-wise non-linear activation function. Note that the last layer has no activation function. 
Parameters $\vtheta$ are updated during the training by minimizing an empirical risk/loss function characterizing the difference between the DNN outputs and the correct outputs, e.g., $R_{S}(\vtheta)=\Sigma_{i=1}^{n}\norm{f_{\vtheta}(\vx_i)-\vy_i)}^2/2n$ for the loss of training dataset $S$, with gradient-based algorithms.
%In the research of deep learning, majority of mathematical researchers insist on fully rigorous proofs \citep{saxe_exact_2013,saxe_information_2019,lampinen_analytic_2019,engel_statistical_2001,aubin_committee_2018,choromanska_loss_2015,mei_mean_2018,rotskoff_parameters_2018,chizat_global_2018,sirignano_mean_2020,jacot_neural_2018,lee_wide_2019}, while majority of industrial researchers aim to deliver end-to-end data-driven progress \citep{lecun_deep_2015}. 
Due to the highly nonlinear nature of the neural network (NN) model, many key theoretical questions raised by Leo Breiman decades ago remain unanswered \citep{breiman1995reflections,zdeborova2020understanding}. This work focuses on one of them---why heavily parameterized neural networks do not overfit the data, which is further backed by recent experimental works in large datasets and deep networks \citep{zhang_understanding_2017}. Note that, establishing a good theoretical understanding of this non-overfitting puzzle has become more and more crucial for the application of DNNs because modern DNN architectures with tons of parameters, e.g., $\sim 10^8$ for VGG19 \citep{simonyan2014very}, $\sim 10^{11}$ for GPT-3 \cite{brown2020language}, indeed achieve huge success in practice.
However, theoretical understanding to this puzzle is not obvious at all, because it contradicts the doctrine in physics and statistical learning theory implied by von Neumann's famous quote ``with four parameters I can fit an elephant" \cite{dyson_meeting_2004}. Existing theories based on idealized models of DNNs, e.g., deep linear network \citep{saxe_exact_2013,saxe_information_2019,lampinen_analytic_2019}, committee machine \citep{engel_statistical_2001,aubin_committee_2018}, spin glass model \citep{choromanska_loss_2015}, mean-field model \citep{mei_mean_2018,rotskoff_parameters_2018,chizat_global_2018,sirignano_mean_2020}, neural tangent kernel \citep{jacot_neural_2018,lee_wide_2019}, which emphasize on fully rigorous mathematical proofs, have difficulties in providing a satisfactory explanation \cite{zdeborova2020understanding}.  

\begin{figure}
\includegraphics[scale=0.4]{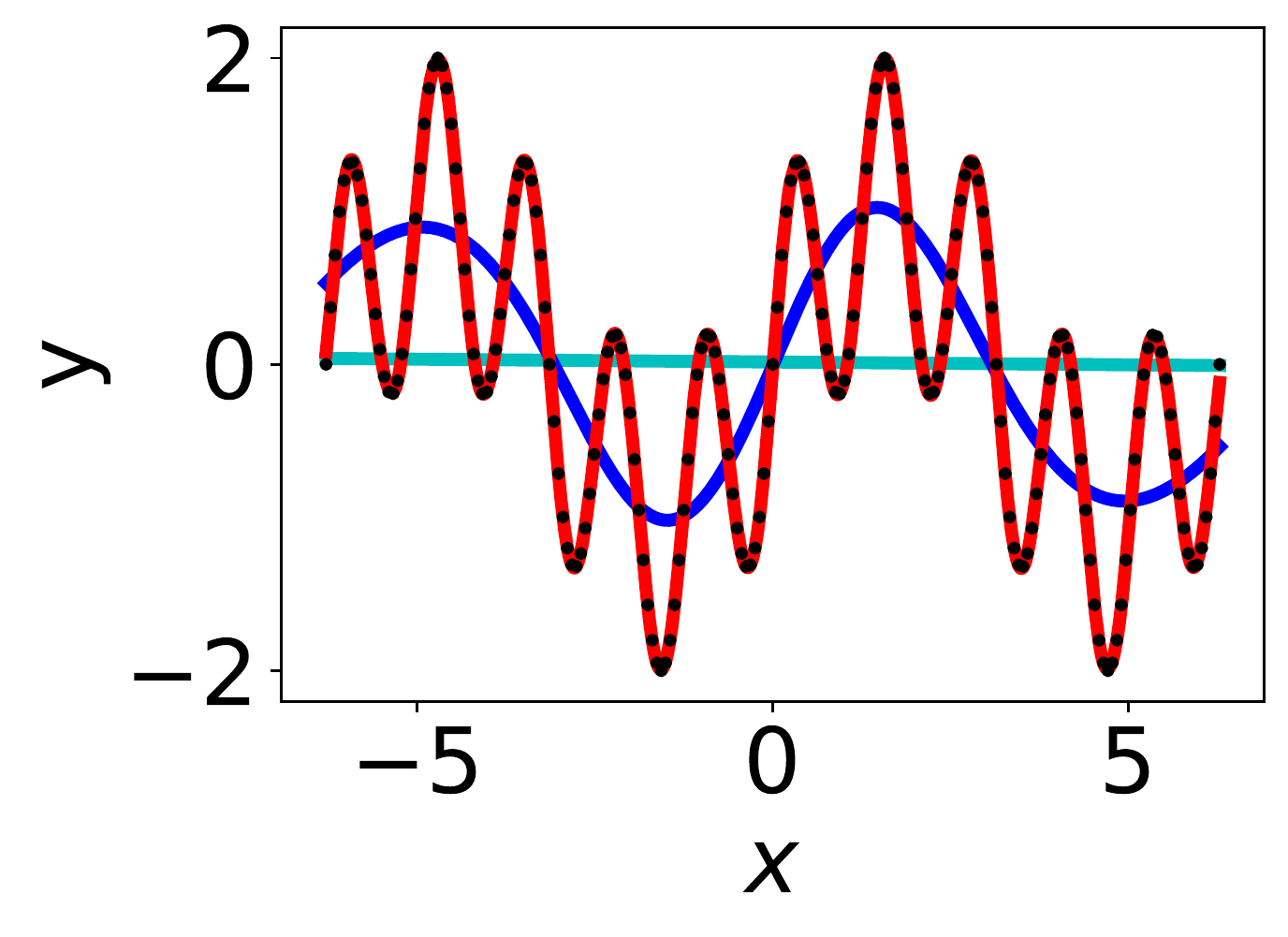}
\caption{\label{fig:1dfp}Illustration of the training process of a DNN. Black dots are training data sampled from target function $\sin(x)+\sin(5x)$. Cyan, blue and red curves indicates $f_{\vtheta(t)}(\vx)$ at training epochs $t=0,2000,17000$, respectively.}
\end{figure}

The training process of NN under gradient flow can be viewed as collective dynamics of a large group of interacting neurons/parameters driven by training data. In analogy to statistical mechanics, there are microscopic levels of NNs caring about the detailed dynamics of each component of $\vtheta$ and macroscopic level caring about dynamics of statistical quantities of $\vtheta$, among which $f_{\vtheta}(\vx)$ as a function-valued quantity is the most important one. At the macroscopic level, it was suggested recently that DNNs learn simple patterns (e.g., certain coarse description or landscape of dataset)  first \citep{arpit2017closer,kalimeris_sgd_2019,valle2018deep}. Based on the intuition that low frequency functions, i.e., functions with energy mainly concentrated at low frequencies, are of low complexity, Refs. \citep{xu_training_2019,xu_frequency_2019,rahaman_spectral_2019,basri2019convergence} quantify the complexity of $f_{\vtheta}(\vx)$ by its frequency composition, and demonstrated the general phenomenon of frequency principle (F-Principle)---NNs often learn low frequencies first. For example, when a DNN is used to fit data generated from $1$-d target function $\sin(x)+\sin(5x)$, while $\sin(x)$ and $\sin(5x)$ have the same amplitude, the low frequency $\sin(x)$ is first captured, and later the target as shown in Fig. \ref{fig:1dfp}. This phenomenon can be robustly observed no matter how overparameterized NNs are. The F-Principle has initiated a series of subsequent works  \cite{rabinowitz_meta-learners_2019,jagtap_adaptive_2020,ronen_convergence_2019,yang_fine-grained_2020,cao2019towards}, and inspired the design of DNN-based algorithms \cite{cai_phase_2019,biland_frequency-aware_2019,liu2019multi,li2019multi,wang2019multi}. 

In this letter, starting from this key macroscopic dynamical feature of F-Principle, we establish a theory for the non-overfitting puzzle.  We propose a linear frequency principle (LFP) model for the phenomenological characterization of the F-Principle. Based on the LFP model, we establish a theory which explains the non-overfitting puzzle, and experimentally test its qualitative predictions about failures of NNs. Furthermore, through an ideal example of two-layer NN in the infinite proper width limit, we unravel how microscopic reality of NN training dynamics statistically gives rise to a LFP model. Finally, we demonstrate the quantitative prediction power of the LFP model through experiments. 

\textit{LFP model.} F-Principle is ``opposite'' to common physical processes with diffusion, in which high frequency modes dissipate faster than low frequency ones. To phenomenologically model such a process, we consider a dynamics in frequency domain, in which each frequency mode evolves to certain target determined by training data $\{\vx_i\in \mathbb{R}^d,f^*(\vx_i)\in \mathbb{R}\}_{i=1}^n$  with a positive rate $\gamma(\vxi)$ decaying as frequency $\vxi\in\mathbb{R}^d$ increases. In particular, we show in this letter that for a wide two-layer NN, explicit form of $\gamma(\vxi)$, which is a linear combination of $\frac{1}{\norm{\vxi}^{d+1}}$ and $\frac{1}{\norm{\vxi}^{d+3}}$, can be derived to accurately predict the NN outputs after training. Before that, we begin with proposing the following  general model for F-Principle,
\begin{equation}
    \partial_{t}\hat{h}(\vxi,t)=-\gamma(\vxi)\left(\hat{h_{\rho}}(\vxi,t)-\hat{f^*_{\rho}}(\vxi)\right), \label{LFPtype}
\end{equation}
where $h(\vx,t)$ models $f_{\vtheta(t)}(\vx)$ with microscopic details neglected, $\hat{(\cdot)}(\vxi)=\int_{\sR^{d}} (\cdot)(\vx)\E^{-\I \vx\cdot \vxi}\diff{\vx}$ is the Fourier transform. The initial condition is set to $h(\vx,0)=h_{\rm ini}(\vx)$. $(\cdot)_{\rho}(\vx) = (\cdot)(\vx)\rho(\vx)$. $\rho(\vx)$ is the data distribution, which can be a continuous function or a probability function for discrete training data points, that is, $\rho(\vx)=\Sigma_{i=1}^{n}\delta(\vx-\vx_i)/n$ ($\delta(\cdot)$ is the dirac delta function), an uncommon part of this dynamics. Since the steady state requires the model prediction equal to the target function only at the empirical training data points, at the steady state, no explicit constraint is imposed on the unseen data points, therefore,  $h(\vx,\infty)$ can drastically deviate from the target $f^*(\vx)$ at unseen data points. We call model (\ref{LFPtype}) \emph{Linear frequency principle (LFP) model}, in which ``linear'' refers to the fact that model (\ref{LFPtype}) is a linear differential equation in $h$. For simplicity, we set $u(\vx,t)=h(\vx,t)-f^*(\vx)$ with the LFP model simplified to $\partial_{t}\hat{u}(\vxi,t)=-\gamma(\vxi)\hat{u_{\rho}}(\vxi,t)$. 

We relate dynamics of model (\ref{LFPtype}) with the dynamics of least square loss, that is,  model (\ref{LFPtype}) is a dissipative process with a decreasing loss (in analogy to energy)
 \begin{equation}
 R_{S}=\frac{1}{2}\int u^2\rho\diff{\vx}=\frac{1}{2}\int \hat{u}\hat{u_\rho}^*\diff{\vxi}.     \label{loss}
 \end{equation}
governed by $\frac{d}{dt}R_{S}=-\int \gamma|\hat{u_\rho}|^2\diff{\vxi}<0$. The dissipation of loss at each frequency is governed by $\partial_{t}(\hat{u}\hat{u_\rho}^*/2)=-\gamma|\hat{u_\rho}|^2$. More importantly, because $\gamma(\vxi)$ is a decaying function by F-Principle, e.g., $\gamma(\vxi)=\frac{1}{\norm{\vxi}^{d+1}}$, loss decreases faster over lower frequencies. This behavior is essential for overcoming the singularity
in $u_{\rho}$ as a summation of delta functions. Otherwise, if $\gamma(\vxi) = \norm{\vxi}^2$, model (\ref{LFPtype}) becomes a heat-diffusion-type equation $\partial_{t} u=-\Delta u_\rho$ in spatial domain and is not well-posed for non-differentiable $u_{\rho}$. As the study of waves by mode decomposition, the coefficient $\gamma(\vxi)$ as a function of frequency $\vxi$  plays an important role in governing the macroscopic phenomenon of training dynamics. For example, for a $\gamma(\vxi)$ decaying with $\vxi$, model (1) first learns the landscape or a simple pattern of the training data, followed by more details or complex patterns, exemplified by the case in Fig. 1. However, for a $\gamma(\vxi)$ increasing with $\vxi$, the learning behavior is opposite. Note that, there are infinite feasible decay functions of $\gamma(\vxi)$ obeying the F-Principle. In general, power-law decay is relevant to an activation of singularity in derivatives, e.g., ReLU, whereas exponential decay is relevant to a smooth activation, e.g., tanh. 

In the following, we further analyze our proposed LFP model (\ref{LFPtype}) in two folds. First, we show that the long-time solution of model (\ref{LFPtype}) is equivalent to the solution of an optimization problem, which reveals the low-frequency bias of the LFP model. Based on the optimization problem, we obtain an generalization error estimate for understanding the non-overfitting puzzle. Second, as an example, we exactly  compute the LFP model for two-layer wide ReLU networks.

\textit{Theory for the non-overfitting puzzle.} Our LFP model searches for the fitting of $n$ points of $f^{*}$ in an infinite dimensional function space. Clearly, it possesses infinite steady states (minimizers of $E$) that satisfy $h(\vx_i)=f^{*}(\vx_i)$ for $i=1,\cdots,n$. If we arbitrarily pick one steady state of $h$, it is likely to generalize poorly, i.e., $h$ deviates drastically from target $f^*$ on unobserved positions, resulting in overfitting as commonly expected from an overparameterized model. However, given proper $h_{\mathrm{ini}}(\vx)$ and $\gamma(\vxi)$, we obtain a unique steady state $h(\vx,\infty)$ (denoted by $h_{\infty}(\vx)$ for simplicity). Exploiting the linearity of the LFP model, we derive that $h_{\infty}(\vx)$ satisfies the following constrained minimization problem
\begin{equation}\label{eq:ndopti}
\begin{aligned}
 \min_{h} &\int \gamma(\vxi)^{-1}|\hat{h}(\vxi)-\hat{h}_{\rm ini}(\vxi)|^2\diff{\vxi},\\
{\rm s.t.} \; &h(\vx_{i})=f^{*}(\vx_i),\;i=1,\ldots,n.
\end{aligned}
\end{equation}
Note that, solution to this problem may generalize poorly if $h_{\mathrm{ini}}$ attains any function. For example, if $h_{\mathrm{ini}}$ attains a ``bad" steady state, then solution of the problem $h=h_{\mathrm{ini}}$ is also ``bad". However, in practice, common initialization of NNs yields small output. Without loss of generality, we consider in the following an unbiased initial function $h_{\rm ini}=0$, which can be achieved in NNs by applying the AntiSymmetrical Initialization (ASI) trick \citep{zhang_type_2019}.

This static minimization problem defines an FP-energy $E_{\gamma}(h)=\int \gamma^{-1}|\hat{h}|^2\diff{\vxi}$ that quantifies the preference of the LFP model among all its steady states. Because $\gamma(\vxi)^{-1}$ is an increasing function, say $\gamma(\vxi)^{-1}=\norm{\vxi}^{d+1}$, the FP-energy $\int \norm{\vxi}^{d+1}|\hat{h}|^2\diff{\vxi}$ amplifies the high frequencies while diminishing low frequencies. By minimizing $E_{\gamma}(h)$, problem (\ref{eq:ndopti}) gives rise to a low frequency fitting, instead of an arbitrary one, of training data. By intuition, if target $f^*$ is indeed low frequency dominant, then $h_{\infty}$ likely well approximates $f^*$ at unobserved positions.

To theoretically demonstrate above intuition, we derive in the following an estimate of the generalization error of $h_{\infty}$ using the {\it a priori} error estimate technique \cite{e_priori_2019}. Because $h(\vx)=f^{*}(\vx)$ is a viable steady state, $E_{\gamma}(h_{\infty})\leq E_{\gamma}(f^*)$ by the minimization problem. Using this constraint on $h_{\infty}$, we obtain that, with probability of at least $1-\delta$,
\begin{equation}
    \Exp_{\vx} (h_{\infty}(\vx)-f^{*}(\vx))^2
    \leq \frac{E_{\gamma}(f^*)}{\sqrt{n}}C_{\gamma}
    \left(2+4\sqrt{2\log(4/\delta)}\right),\label{generror}
\end{equation}
where $C_{\gamma}$ is a constant depending on $\gamma$. Error reduces with more training data as expected with a decay rate $1/\sqrt{n}$ similar to Monte-Carlo method. Importantly, because $E_{\gamma}(f^*)$ strongly amplifies high frequencies of $f^*$, the more high-frequency components the target function $f^*$ possesses, the worse $h_{\infty}$ may generalize.

Above theory explains the non-overfiting puzzle of NNs as follows: regardless of the number of parameters of NNs,  the F-Principle dynamics finds for an overparameterized NN a low frequency fitting of training data, which unlikely overfits a low frequency target function (since the FP-norm is small for low-frequency function). Specifically, it predicts the following qualitative behaviors of NNs.\\
(i) {\it Preference.} NNs preferentially learn low frequency fittings of training data; \\
(ii) {\it Success.} NNs often generalize for low frequency dominant target functions; \\
(iii) {\it Failure.} NNs likely overfit a high frequency target function.

\begin{figure}
\begin{centering}
\subfloat[]{\begin{centering}
\includegraphics[scale=0.28]{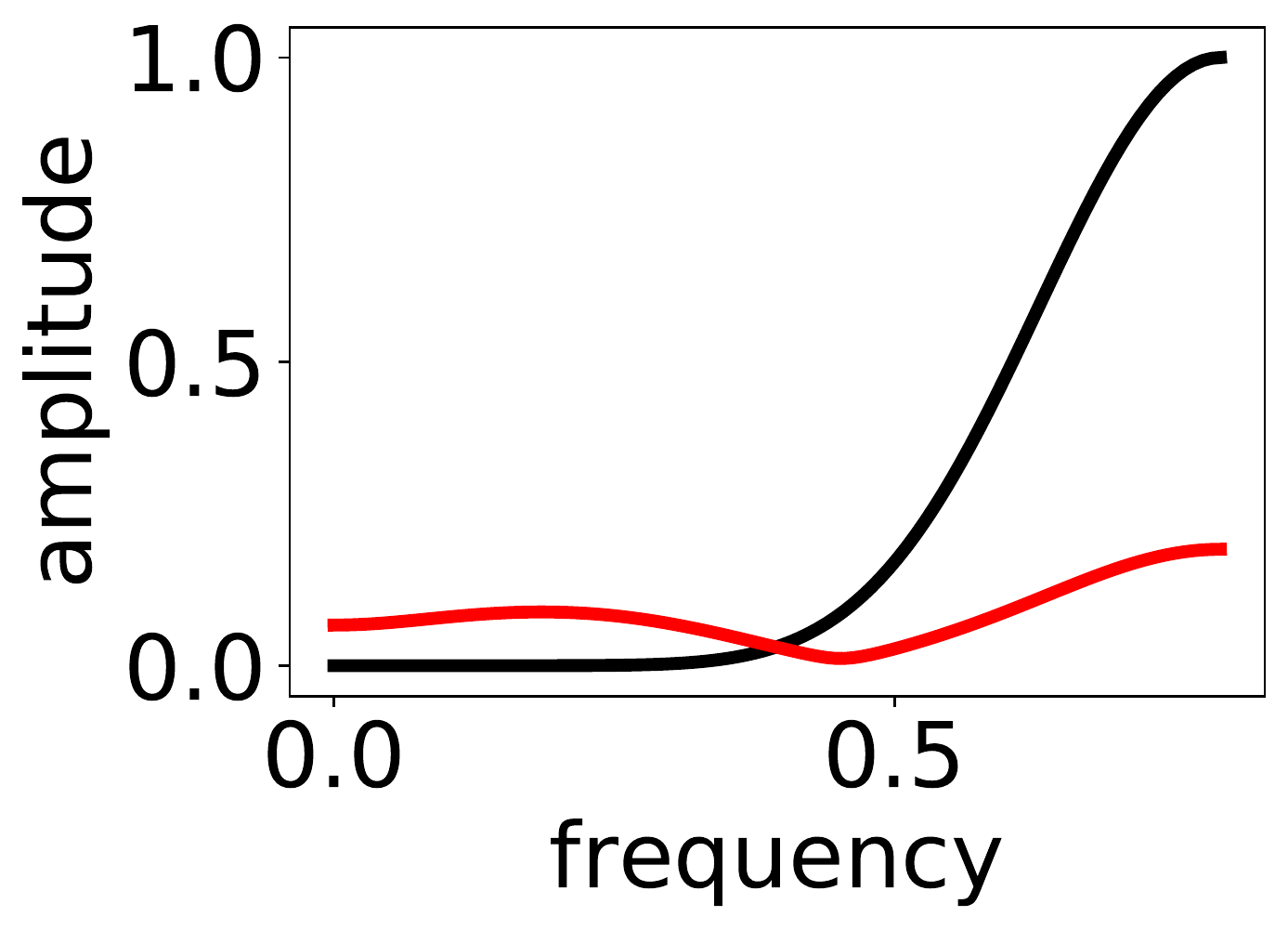} 
\par\end{centering}
}\subfloat[]{\begin{centering}
\includegraphics[scale=0.28]{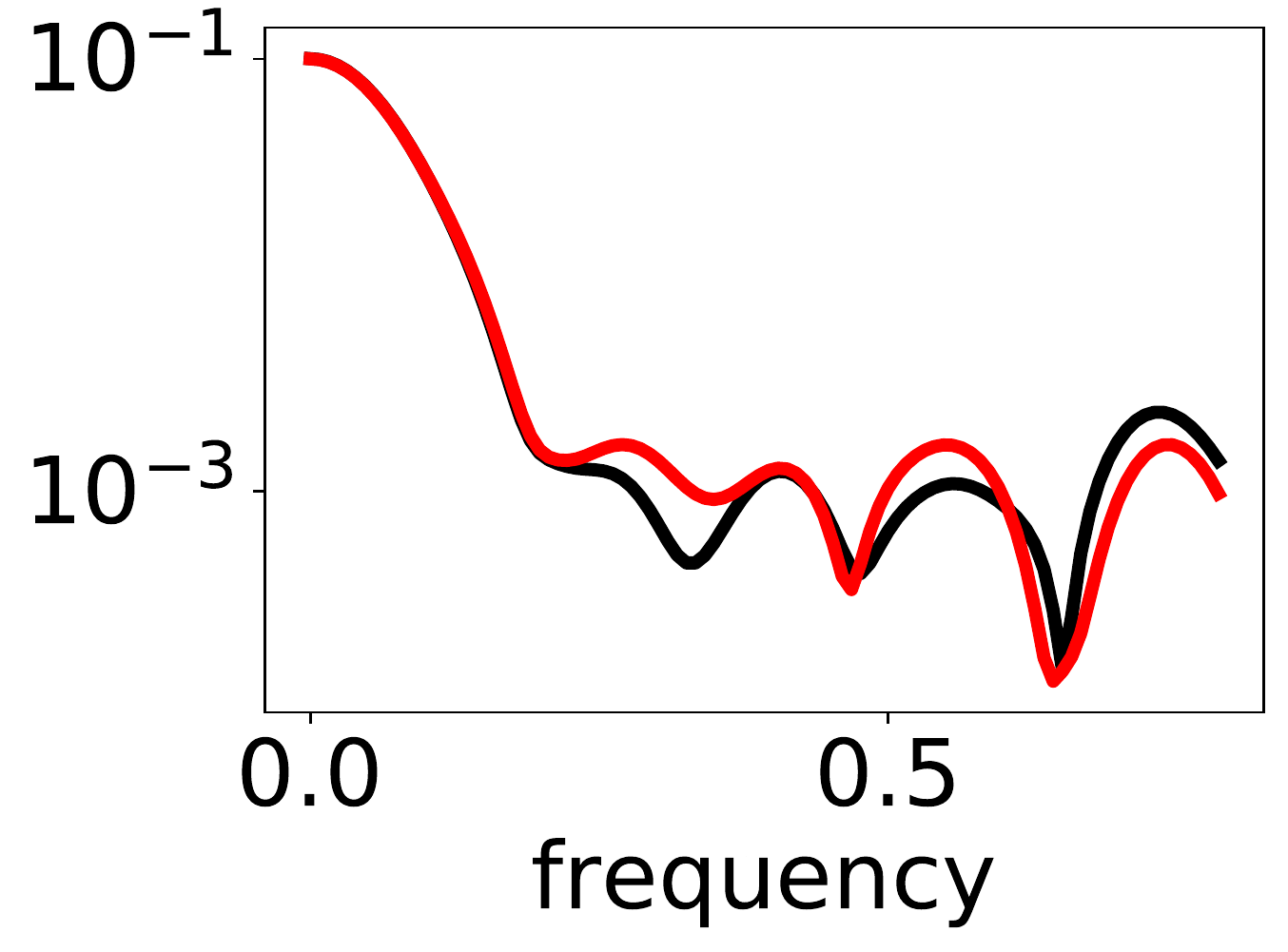} 
\par\end{centering}
}
\par\end{centering}
\caption{Frequency composition (amplitude vs. frequency) of target (black) and the well-trained DNN (red) along the first principle component direction of inputs of training data.  (a) Target: $10$-d parity function; NN: three-layer fully connected net. (b) Target: CIFAR10; NN: two convolutional layers with a fully-connected layer. \label{fig:cmft} }
\end{figure}

In the following, we test whether these predictions well hold for NNs in experiments. In the first experiment, we use a DNN to fit high dimensional high frequency dominant data sampled from a parity function   $f(\vec{x})=\prod_{j=1}^d x_{j}$ defined on $\Omega=\{-1,1\}^{d}$, whose Fourier transform $(-\I)^{d}\prod_{j=1}^d\sin2\pi k_{j}$ for $\vk \in[-\frac{1}{4},\frac{1}{4}]^{d}$ peaks at its highest frequencies $\vk \in\{-\frac{1}{4},\frac{1}{4}\}^{d}$. The difficulty of learning the parity function with NNs is well-known \cite{minsky2017perceptrons,allender1996circuit}. We provide a frequency perspective to understand this learning difficulty. For high-dimensional function, we perform a non-uniform discrete Fourier transform on the first principle direction of a training data set. As demonstrated in Fig. \ref{fig:cmft}a, the well-trained DNN indeed preferentially learns more low frequencies and less high frequencies comparing to the target. Furthermore, as model predicted, the DNN generalizes badly with a low test accuracy $38\%$ no more than chance-level $50\%$ (while training accuracy is $100\%$!). In the second experiment, we use the widely considered image classification dataset of CIFAR10 as an example, on which a well-trained DNN achieves a test accuracy $68\%$ much higher than chance-level $10\%$, and compute its frequency composition by non-uniform discrete Fourier transform. As shown in Fig. \ref{fig:cmft}b, the target is indeed dominated by low frequencies. Actually, this low frequency dominance property for most real high dimensional image data can be intuitively understood based on the common sense that a small perturbation in input image mostly does not change the its category as output. Furthermore, as predicted, DNN preferentially learns the low frequencies better than the high ones, leading to a good generalization.

\textit{LFP model derived from a two-layer NN.}
Analysis of the training process of a multi-layer ($L\geq2$) NN is well-known difficult \cite{zdeborova2020understanding}. Recently, based on a dynamical regime of neural tangent kernel (NTK), where the gradient flow of overparameterized NNs can be effectively linearized around initialization, fruitful mathematical theorems were proved at an abstract level about the behavior of NNs \cite{jacot_neural_2018,arora2019fine,e2020comparative}. Still, deriving explicitly the linearized dynamics of even a two-layer NN, which already possesses similar nontrivial training and generalization behavior as deeper NNs, for quantitative analysis is a challenging task.
In this part, we present such a derivation in frequency domain, which yields a LFP model with a specific $\gamma(\vxi)$ depending on detailed setups of the target NN, such as smoothness of $\sigma$ and statistics of $\vtheta(0)$.  Note that, since the F-Principle generally exists in deeper NNs and in both NTK and non-NTK regimes, the mechanism unraveled by the above theoretical analysis, i.e., low frequency first learning dynamics leads to a low frequency fitting of data, applies to general DNNs where the NTK theory can drastically fail.

% We study the LFP model of two-layer NNs in NTK regime because NNs are linear model with respect to training parameters in NTK regime, which greatly alleviate the technical complexity.} However, note that since the F-Principle generally exists in non-NTK regime, our proposed model (\ref{LFPtype}) is not a simple Fourier version of NTK dynamics. In addition, the error estimate in (\ref{generror}) is also valid for general model (\ref{LFPtype}).}

Considering the following two-layer neural network 
\begin{equation}
f(\vx;\vtheta)=\frac{1}{\sqrt{m}}\sum_{j=1}^m a_j\sigma(\vw_j^{\T}\vx+r_j c_j),\label{eq:2layerNN}
\end{equation}
where   $r_j:=\abs{\vw_j}$ and $\sigma(x)=\max(0,x)$,
i.e., the widely used ReLU (rectified linear unit) activation. Note that our following derivation applies similarly to other $\sigma$ such as sigmoid or tanh activation. Denote $\vp_j=(a_j,\vw_j,c_j)^\T\in\sR^{d+2}$, $\vtheta=(\vp_1,\cdots,\vp_m)$. During the learning process, i.e., fitting training data $\{(\vx_i,f^{*}(\vx_i))\}_{i=1}^{n}$ generated from a target function $f^{*}(\vx)$ by Model (\ref{eq:2layerNN}), $\vtheta$ evolves
by the gradient descent with dynamics at continuous limit
\begin{equation}
\dot{\vtheta}=-\nabla R_{S}(\vtheta),\label{eq:gf}
\end{equation}
with mean-squared error (MSE) loss $R_{S}(\vtheta)$ of the empirical sample distribution in Eq. (\ref{loss}).  

%\begin{equation}
%R(\vtheta)=\frac{1}{2}\int_{\mathbb{R}^{d}}(f(\vx;\vtheta)-f^{*}(\vx))^{2}\rho(\vx)\D \vx,\label{eq:loss}
%\end{equation}
%where $\rho(\vx)=\frac{1}{n}\Sigma_{i=1}^{n}\delta(\vx-\vx_i)$  is the empirical sample distribution.

At initialization, $a_j,\vw_j,c_j$ for $j=1,\cdots,m$ are sampled independently from random distributions under mild assumptions that (i) distribution of $\vw_j/\abs{\vw_j}$ is uniform on the unit sphere; (ii) variance of $c_j$, denoted by $\sigma_c^2$, is sufficiently large.

In general, dynamics  (\ref{eq:gf}) is difficult to be analyzed due to its high-dimensional and highly nonlinear nature similar to particle systems in statistical mechanics \cite{rotskoff_parameters_2018}. 
In the following, we show how a LFP macroscopic statistical description of above dynamics can be derived at the infinite neuron limit $m\to \infty$, which has been considered in Refs. \citep{jacot_neural_2018,lee_wide_2019,rotskoff_parameters_2018,mei_mean_2018,sirignano_mean_2020}, in analogy to the thermodynamic limit.
This limit with the scaling factor of $1/\sqrt{m}$  in NN (\ref{eq:2layerNN}) makes its linearization around initialization 
\begin{equation}
f^{{\rm lin}}\left(\vx;\vtheta(t)\right)=f\left(\vx;\vtheta(0)\right)+\nabla_{\vtheta}f\left(\vx;\vtheta(0)\right)\left(\vtheta(t)-\vtheta(0)\right) \label{eq:linearNN}
\end{equation}
an effective approximation of $f\left(\vx;\vtheta(t)\right)$, i.e.,  $f^{{\rm lin}}\left(\vx;\vtheta(t)\right)\approx f\left(\vx;\vtheta(t)\right)$ for any $t$, as demonstrated by both theoretical and empirical studies of neural tangent kernels (NTK) \citep{jacot_neural_2018,lee_wide_2019}.  Note that,  $f^{{\rm lin}}\left(\vx;\vtheta(t)\right)$, linear in $\vtheta$ and nonlinear in $\vx$, reserves the universal approximation power of $f\left(\vx;\vtheta(t)\right)$ at $m\to \infty$. In the following, we do not distinguish $f\left(\vx;\vtheta(t)\right)$ from $f^{{\rm lin}}\left(\vx;\vtheta(t)\right)$.

Again, analogous to statistical mechanics, while Dynamics (\ref{eq:gf}) act at a microscopic level on parameters of each neuron, function $f\left(\vx,\vtheta(t)\right)$ for the fitting problem is macroscopic.  For simplicity, we denote $f(\vx,\vtheta(t))$ by $f(\vx,t)$.  The evolution of  $f\left(\vx,t\right)$  in the NTK regime follows gradient flow, i.e.,
\begin{align*}
     \partial_t  f\left(\vx,t\right)=\nabla_{\vtheta}f\left(\vx,t\right)\cdot \partial_t \vtheta=-\int_{\mathbb{R}}u_{\rho}(\vx^\prime)K_{\vtheta}(\vx,\vx^{\prime}) \D \vx^{\prime},
\end{align*}
where $K_{\vtheta}(\vx,\vx^\prime)=\nabla_{\vtheta}f\left(\vx^{\prime},\vtheta(0)\right)\nabla_{\vtheta}f\left(\vx,\vtheta(0)\right)$, $u(\vx)=f(\vx,t)-f^{*}(\vx)$, $u_{\rho}(\vx)=u(\vx)\rho(\vx)$. This gradient flow applies for deep neural networks with arbitrary hidden layers in the NTK regime. However, to derive an explicit form of the kernel $K_{\vtheta}$, we limit our analysis to the two-layer ReLU neural network. 
By applying Fourier transform to both side of above equation, we obtain with approximation 
\begin{equation}\label{eq:LFP_NN}
\partial_{t}\hat{u}(\vxi,t)=-\left[\frac{\left\langle a^{2}+r^{2}\right\rangle _{a,r}}{\norm{\vxi}^{d+3}}+\frac{\left\langle a^{2}r^{2}\right\rangle _{a,r}}{\norm{\vxi}^{d+1}}\right]\hat{u_{\rho}}(\vxi,t),
\end{equation}
where $\left\langle \cdot \right\rangle _{a,r}$ is the expectation with respect to the initial distribution of $a$ and $r$. Clearly, it is a LFP model prioritizes the learning of low frequencies quantified by mixed power law decay. This power law decay results from the decay of the spectrum of $\sigma$ depending on its smoothness. For a sigmoid or tanh activation, an exponential decay will be obtained. This model signifies the analogy between NN and statistical mechanics that the learning process of NN with a large number of neurons is effectively captured by several statistics, e.g., $\left\langle a^{2}+r^{2}\right\rangle _{a,r}$ and $\left\langle a^{2}r^{2}\right\rangle _{a,r}$, with microscopic details neglected. 
Remark that, to derive Model (\ref{eq:LFP_NN}), we ignore an additional term arising from the rotation of $\vw$'s for $d\geq2$, that is, $\frac{\left\langle r^{2}\right\rangle _{r}}{\left\Vert \vxi\right\Vert _{2}^{d+1}}\Delta_{\vxi^{\bot}}\hat{u_{\rho}}(\vxi)$, where $\Delta_{\vxi^{\bot}}$ indicates a Laplacian at the subspace orthogonal to $\vxi$. 
While F-Principle always holds due to the power-law decay, there is mild extra effect results from this term in practice. Details about such an effect remains a problem for future study. This suggests a wider class of \emph{generalized LFP model}, in which $\gamma(\vxi)$ can be a general linear operator in frequency domain. Detailed properties about the generalized model remains a problem for future study.

To analyze Model (\ref{eq:LFP_NN}), we resort to its equivalent optimization problem as discussed before. Based on the equivalent optimization problem in (\ref{eq:ndopti}) and the error estimate in (\ref{generror}), the non-overfitting puzzle for two-layer wide ReLU NNs can be explained. Next, we analyze each decaying term for 1-d problems ($d=1$). When $1/\xi^2$ term dominates, the corresponding minimization problem Eq. (\ref{eq:ndopti}) rewritten into spatial domain yields 
\begin{equation}\label{eq:LS}
\begin{aligned}
 &\min_{h} \int|h^{'}(x)-h_{\rm ini}^{'}(x)|^2\diff{x},\\
{\rm s.t.} & \quad h(\vx_{i})=f^{*}(\vx_i),i=1,\cdots,n,
\end{aligned}
\end{equation}
where $'$ indicates differentiation. For $h_{\rm ini}(x)=0$, Eq. (\ref{eq:LS}) indicates a linear spline interpolation. Similarly, when $1/\xi^4$ dominates, $\int|h^{''}(x)-h_{\rm ini}^{''}(x)|^2\diff{x}$ is minimized, indicating a cubic spline. In general, above two power law decay coexist, giving rise to a specific mixture of linear and cubic splines. For high dimensional problems, the model prediction is difficult to interpret because the order of differentiation depends on $d$ and can be fractal.

\begin{figure}
\subfloat[]{\begin{centering}
\includegraphics[scale=0.27]{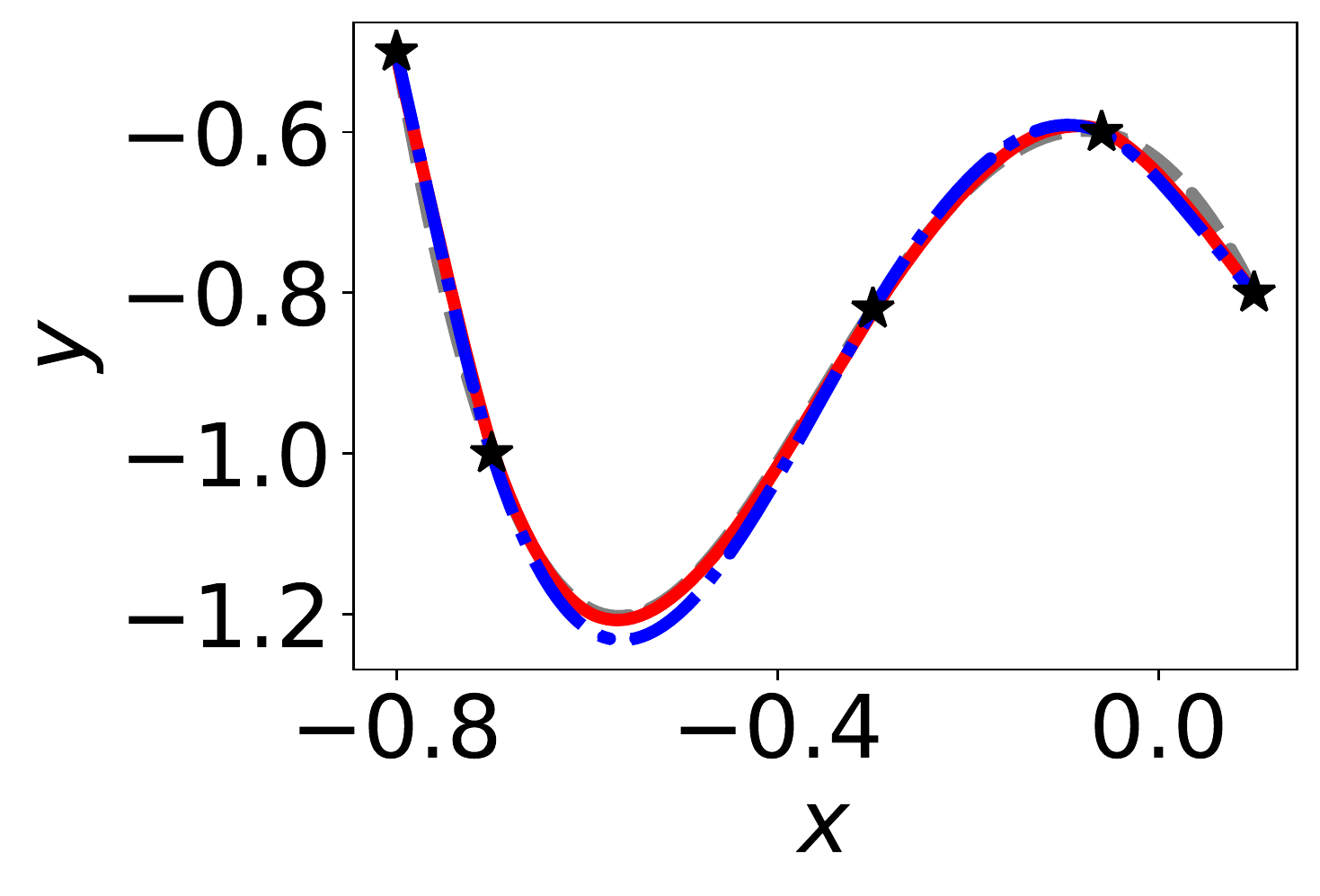} 
\par\end{centering}
}\subfloat[]{\begin{centering}
\includegraphics[scale=0.27]{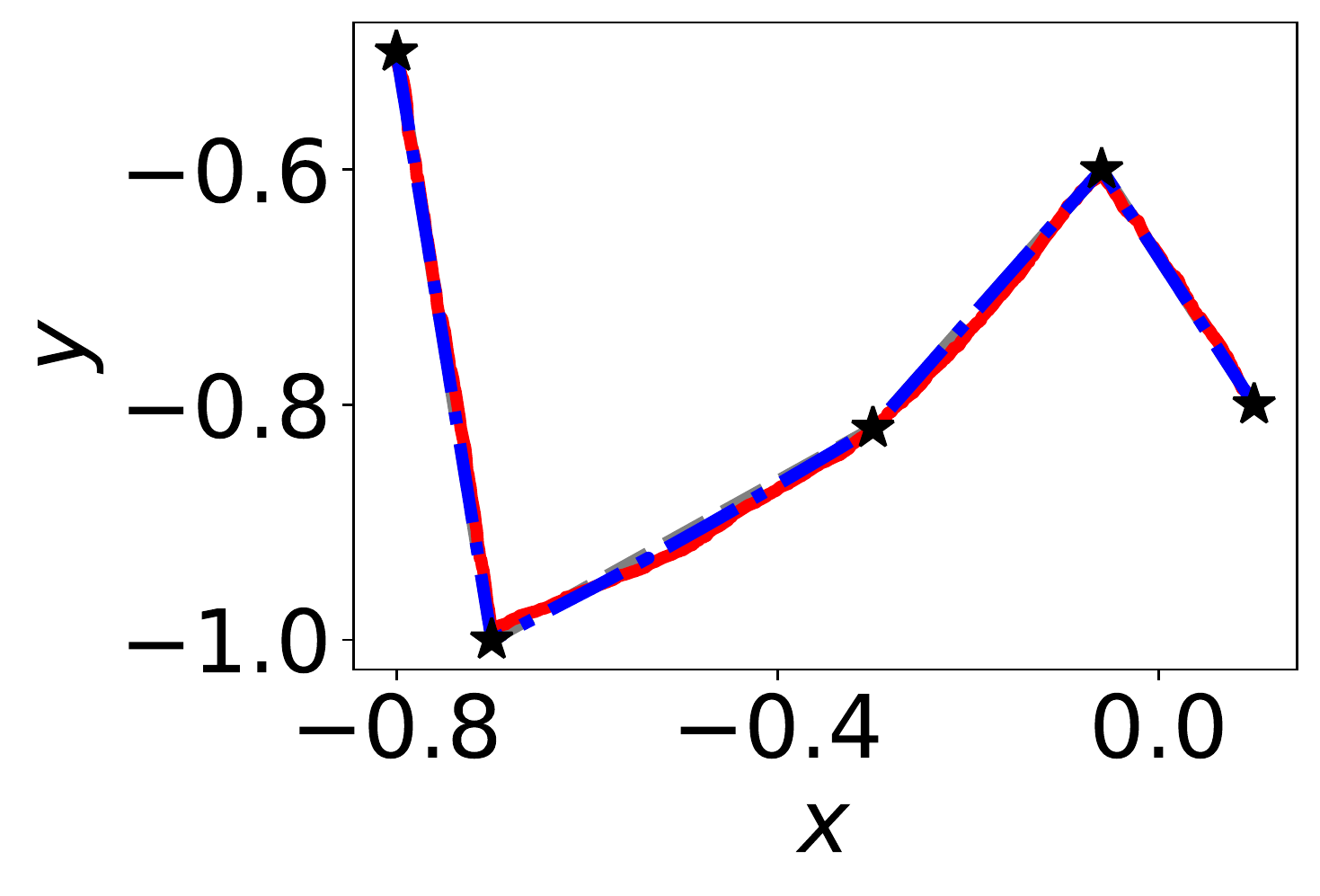} 
\par\end{centering}
}
\caption{$f_{\rm NN}$ (red solid) vs. $f_{\rm LFP}$ (blue dashed dot) vs. splines (grey dashed, cubic spline for (a) and linear spline for (b)) for a $1$-d problem. All curves nearly overlap with one other. Two-layer NN Eq. (\ref{eq:2layerNN}) of $40000$ hidden neurons is initialized with (a) $\left\langle a^{2}+r^{2}\right\rangle _{a,r}\gg \left\langle a^{2}r^{2}\right\rangle _{a,r}$, and (b) $\left\langle a^{2}+r^{2}\right\rangle _{a,r}\ll \left\langle a^{2}r^{2}\right\rangle _{a,r}$. Black stars indicates training data.\label{fig:2relu} }
\end{figure}

\begin{figure}
\subfloat[]{
\includegraphics[scale=0.27]{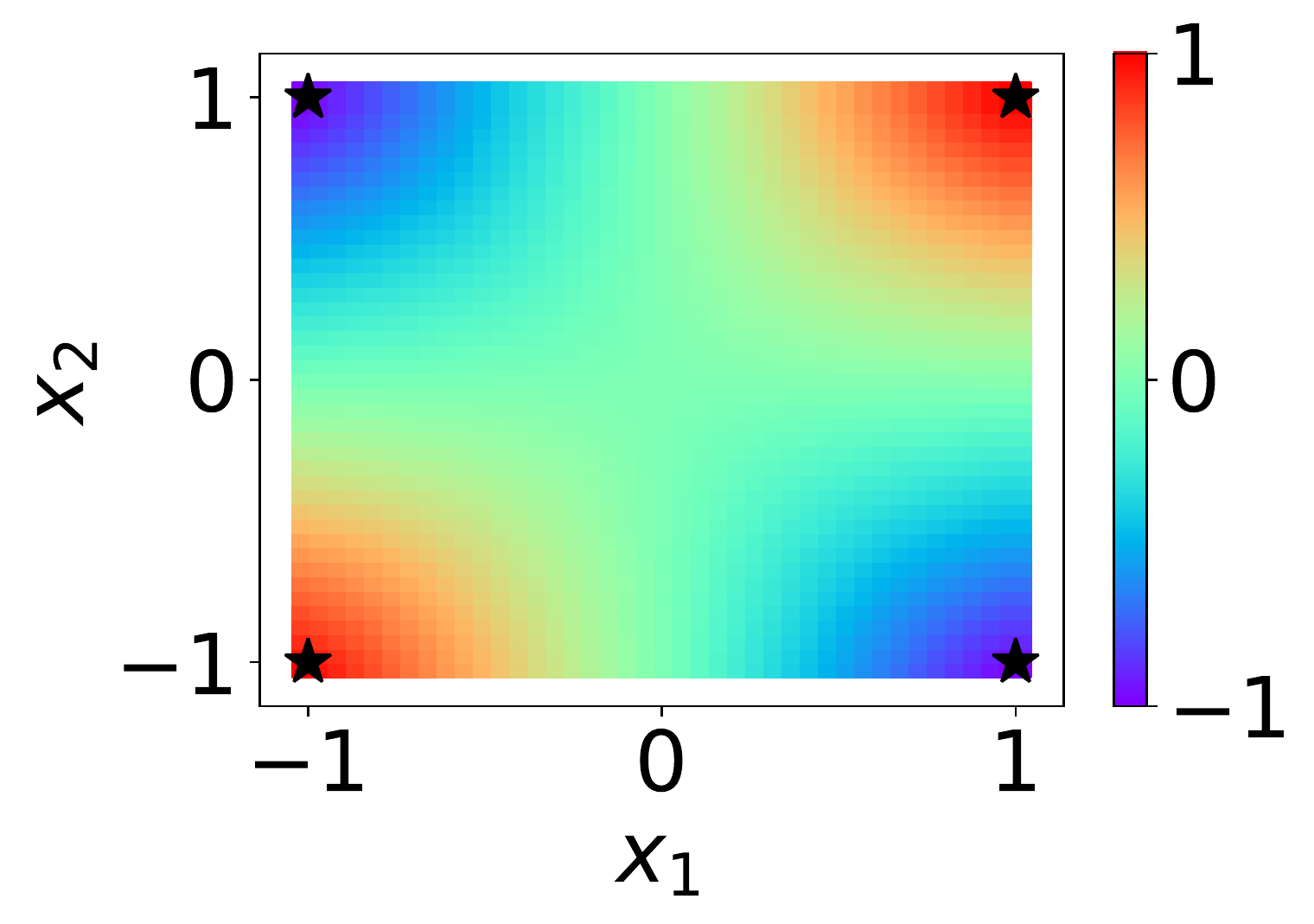} 
}\subfloat[]{
\includegraphics[scale=0.27]{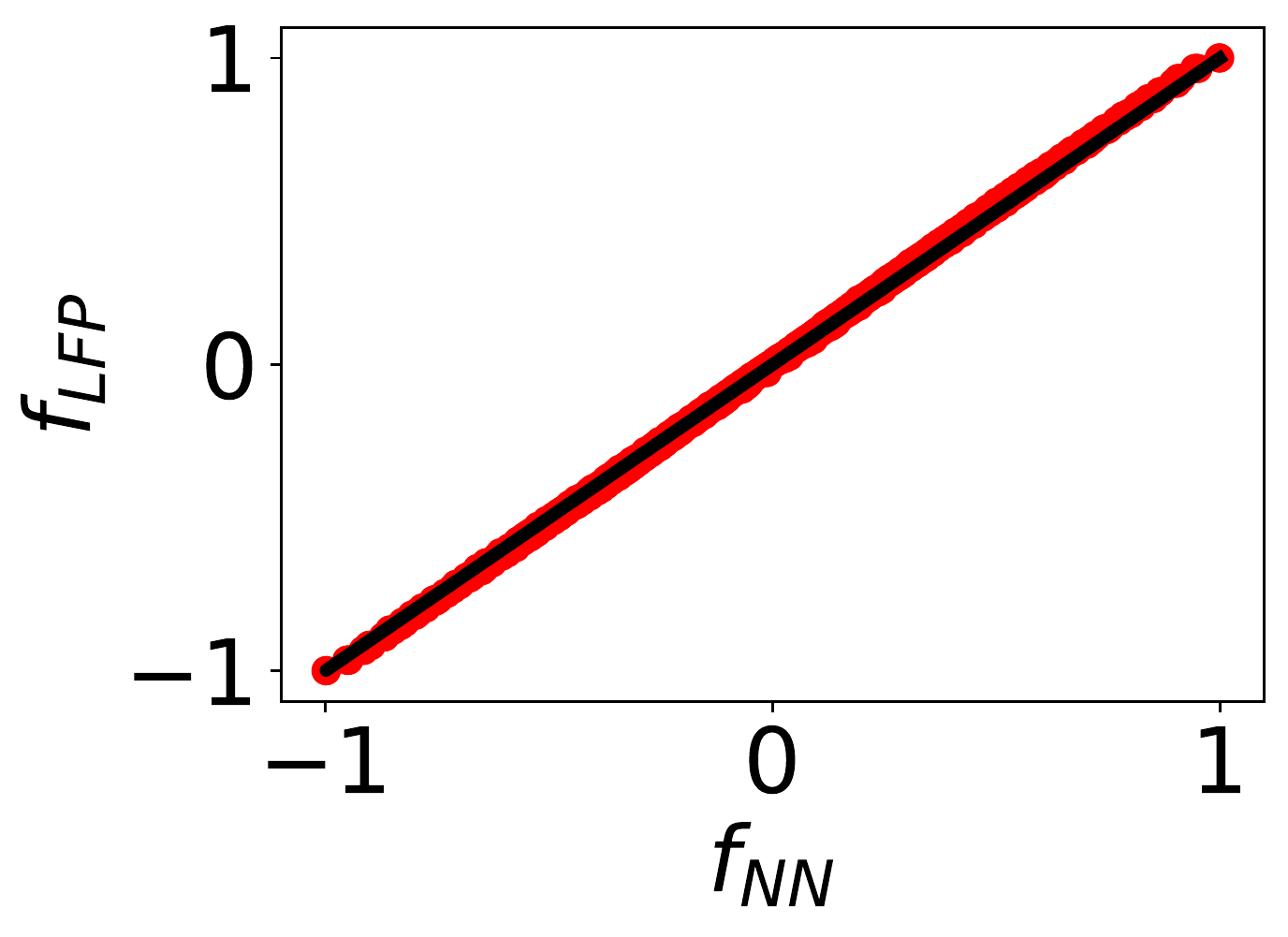} 
}
\caption{$2$-d XOR problem with four training data indicated by black stars learned by a two-layer NN Eq. (\ref{eq:2layerNN}) of $160000$ hidden neurons. (a) $f_{\rm NN}$ illustrated in color scale. (b) $f_{\rm LFP}$ (ordinate) vs. $f_{\rm NN}$ (abscissa) represented by red dots evaluated over whole input domain $[-1,1]^2$. The black line indicates the identity function. \label{fig:2relu-1} }
\end{figure}

In the following, we examine experimentally the quantitative prediction power of LFP model Eq. (\ref{eq:LFP_NN}). For the convenience of notation, solution predicted by the LFP model (\ref{eq:LFP_NN}) is denoted as $f_{\rm LFP}(\vx)=f(\vx,\infty)$. Function learned by NN is denoted by $f_{\rm NN}(\vx)$. As shown in Fig. \ref{fig:2relu}, for a $1$-d problem,  $f_{\rm LFP}(\vx)$ accurately predicts $f_{\rm NN}(\vx)$ over two different initializations. As predicted by above analysis, a wide NN initialized with $\left\langle a^{2}+r^{2}\right\rangle _{a,r}\gg \left\langle a^{2}r^{2}\right\rangle _{a,r}$ learns approximately a  cubic spline, whereas $\left\langle a^{2}+r^{2}\right\rangle _{a,r}\ll \left\langle a^{2}r^{2}\right\rangle _{a,r}$ a linear spline. For $d=2$, we consider the famous XOR problem, which cannot be solved by one-layer neural networks \citep{minsky2017perceptrons}. The training samples consist of four
points represented by black stars in Fig. \ref{fig:2relu-1}a. As
shown in Fig. \ref{fig:2relu-1}b, our LFP model predicts accurately outputs of the well-trained NN over the input domain $[-1,1]^{2}$.

\textit{Discussion.} In this letter, we propose the phenomenological LFP model that explains the absence of overfitting in NNs by its low frequency preference. Our theory informs that NNs are no panacea to all difficult problems and are bad in general for fitting high frequency target functions. As an example, it had been demonstrated that a standard DNN fails drastically for ground state fitting of a frustrated quantum magnet with a rapidly oscillating ground-state characteristic function \cite{cai_approximating_2018}. Therefore, to solve a broad spectrum of problems with practical success, it is important to take into account the low frequency preference of DNNs in the algorithm design. Our work on F-Principle is only a starting point to a more comprehensive understanding of NNs. In the future, the role of depth, width, optimization methods and other hyperparameters in fine tuning the F-Principle dynamics will be studied in detail.
Importantly, more preferences (inductive biases) of NNs, which are keys to open the "black box", need to be unraveled. Specifically, the physics approach from phenomenological study based on carefully designed experiments to theoretical study based on effective models can play an important role as demonstrated by the series of works on F-Principle.

\section{Acknowledgement}
We thank Hugues Chate for critical reading and suggestions on the manuscript. We also thank David W. MacLaughlin, Haijun Zhou, Leihan Tang, Hepeng Zhang, and Yongfeng Zhao for helpful comments on the manuscript. Z.X. is supported by National Key R\&D Program of China (2019YFA0709503), Shanghai Sailing Program, Natural Science Foundation of Shanghai (20ZR1429000), NSFC 62002221 and partially supported by HPC of School of Mathematical Sciences at Shanghai Jiao Tong University.  

% 3 experiment procedure and code share.

\bibliographystyle{apsrev4-1}
\bibliography{LFP}

\begin{thebibliography}{49}%
\makeatletter
\providecommand \@ifxundefined [1]{%
 \@ifx{#1\undefined}
}%
\providecommand \@ifnum [1]{%
 \ifnum #1\expandafter \@firstoftwo
 \else \expandafter \@secondoftwo
 \fi
}%
\providecommand \@ifx [1]{%
 \ifx #1\expandafter \@firstoftwo
 \else \expandafter \@secondoftwo
 \fi
}%
\providecommand \natexlab [1]{#1}%
\providecommand \enquote  [1]{``#1''}%
\providecommand \bibnamefont  [1]{#1}%
\providecommand \bibfnamefont [1]{#1}%
\providecommand \citenamefont [1]{#1}%
\providecommand \href@noop [0]{\@secondoftwo}%
\providecommand \href [0]{\begingroup \@sanitize@url \@href}%
\providecommand \@href[1]{\@@startlink{#1}\@@href}%
\providecommand \@@href[1]{\endgroup#1\@@endlink}%
\providecommand \@sanitize@url [0]{\catcode `\\12\catcode `\$12\catcode
  `\&12\catcode `\#12\catcode `\^12\catcode `\_12\catcode `\%12\relax}%
\providecommand \@@startlink[1]{}%
\providecommand \@@endlink[0]{}%
\providecommand \url  [0]{\begingroup\@sanitize@url \@url }%
\providecommand \@url [1]{\endgroup\@href {#1}{\urlprefix }}%
\providecommand \urlprefix  [0]{URL }%
\providecommand \Eprint [0]{\href }%
\providecommand \doibase [0]{http://dx.doi.org/}%
\providecommand \selectlanguage [0]{\@gobble}%
\providecommand \bibinfo  [0]{\@secondoftwo}%
\providecommand \bibfield  [0]{\@secondoftwo}%
\providecommand \translation [1]{[#1]}%
\providecommand \BibitemOpen [0]{}%
\providecommand \bibitemStop [0]{}%
\providecommand \bibitemNoStop [0]{.\EOS\space}%
\providecommand \EOS [0]{\spacefactor3000\relax}%
\providecommand \BibitemShut  [1]{\csname bibitem#1\endcsname}%
\let\auto@bib@innerbib\@empty
%</preamble>
\bibitem [{\citenamefont {Aurisano}\ \emph {et~al.}(2016)\citenamefont
  {Aurisano}, \citenamefont {Radovic}, \citenamefont {Rocco}, \citenamefont
  {Himmel}, \citenamefont {Messier}, \citenamefont {Niner}, \citenamefont
  {Pawloski}, \citenamefont {Psihas}, \citenamefont {Sousa},\ and\
  \citenamefont {Vahle}}]{aurisano_convolutional_2016}%
  \BibitemOpen
  \bibfield  {author} {\bibinfo {author} {\bibfnamefont {A.}~\bibnamefont
  {Aurisano}}, \bibinfo {author} {\bibfnamefont {A.}~\bibnamefont {Radovic}},
  \bibinfo {author} {\bibfnamefont {D.}~\bibnamefont {Rocco}}, \bibinfo
  {author} {\bibfnamefont {A.}~\bibnamefont {Himmel}}, \bibinfo {author}
  {\bibfnamefont {M.~D.}\ \bibnamefont {Messier}}, \bibinfo {author}
  {\bibfnamefont {E.}~\bibnamefont {Niner}}, \bibinfo {author} {\bibfnamefont
  {G.}~\bibnamefont {Pawloski}}, \bibinfo {author} {\bibfnamefont
  {F.}~\bibnamefont {Psihas}}, \bibinfo {author} {\bibfnamefont
  {A.}~\bibnamefont {Sousa}}, \ and\ \bibinfo {author} {\bibfnamefont
  {P.}~\bibnamefont {Vahle}},\ }\href {\doibase 10.1088/1748-0221/11/09/P09001}
  {\bibfield  {journal} {\bibinfo  {journal} {Journal of Instrumentation}\
  }\textbf {\bibinfo {volume} {11}},\ \bibinfo {pages} {P09001} (\bibinfo
  {year} {2016})}\BibitemShut {NoStop}%
\bibitem [{\citenamefont {Zhang}\ \emph {et~al.}(2018)\citenamefont {Zhang},
  \citenamefont {Han}, \citenamefont {Wang}, \citenamefont {Car},\ and\
  \citenamefont {E}}]{zhang_deep_2018}%
  \BibitemOpen
  \bibfield  {author} {\bibinfo {author} {\bibfnamefont {L.}~\bibnamefont
  {Zhang}}, \bibinfo {author} {\bibfnamefont {J.}~\bibnamefont {Han}}, \bibinfo
  {author} {\bibfnamefont {H.}~\bibnamefont {Wang}}, \bibinfo {author}
  {\bibfnamefont {R.}~\bibnamefont {Car}}, \ and\ \bibinfo {author}
  {\bibfnamefont {W.}~\bibnamefont {E}},\ }\href {\doibase
  10.1103/PhysRevLett.120.143001} {\bibfield  {journal} {\bibinfo  {journal}
  {Physical Review Letters}\ }\textbf {\bibinfo {volume} {120}},\ \bibinfo
  {pages} {143001} (\bibinfo {year} {2018})}\BibitemShut {NoStop}%
\bibitem [{\citenamefont {Guest}\ \emph {et~al.}(2018)\citenamefont {Guest},
  \citenamefont {Cranmer},\ and\ \citenamefont {Whiteson}}]{guest_deep_2018}%
  \BibitemOpen
  \bibfield  {author} {\bibinfo {author} {\bibfnamefont {D.}~\bibnamefont
  {Guest}}, \bibinfo {author} {\bibfnamefont {K.}~\bibnamefont {Cranmer}}, \
  and\ \bibinfo {author} {\bibfnamefont {D.}~\bibnamefont {Whiteson}},\ }\href
  {\doibase 10.1146/annurev-nucl-101917-021019} {\bibfield  {journal} {\bibinfo
   {journal} {Annual Review of Nuclear and Particle Science}\ }\textbf
  {\bibinfo {volume} {68}},\ \bibinfo {pages} {161} (\bibinfo {year}
  {2018})}\BibitemShut {NoStop}%
\bibitem [{\citenamefont {Radovic}\ \emph {et~al.}(2018)\citenamefont
  {Radovic}, \citenamefont {Williams}, \citenamefont {Rousseau}, \citenamefont
  {Kagan}, \citenamefont {Bonacorsi}, \citenamefont {Himmel}, \citenamefont
  {Aurisano}, \citenamefont {Terao},\ and\ \citenamefont
  {Wongjirad}}]{radovic_machine_2018}%
  \BibitemOpen
  \bibfield  {author} {\bibinfo {author} {\bibfnamefont {A.}~\bibnamefont
  {Radovic}}, \bibinfo {author} {\bibfnamefont {M.}~\bibnamefont {Williams}},
  \bibinfo {author} {\bibfnamefont {D.}~\bibnamefont {Rousseau}}, \bibinfo
  {author} {\bibfnamefont {M.}~\bibnamefont {Kagan}}, \bibinfo {author}
  {\bibfnamefont {D.}~\bibnamefont {Bonacorsi}}, \bibinfo {author}
  {\bibfnamefont {A.}~\bibnamefont {Himmel}}, \bibinfo {author} {\bibfnamefont
  {A.}~\bibnamefont {Aurisano}}, \bibinfo {author} {\bibfnamefont
  {K.}~\bibnamefont {Terao}}, \ and\ \bibinfo {author} {\bibfnamefont
  {T.}~\bibnamefont {Wongjirad}},\ }\href {\doibase 10.1038/s41586-018-0361-2}
  {\bibfield  {journal} {\bibinfo  {journal} {Nature}\ }\textbf {\bibinfo
  {volume} {560}},\ \bibinfo {pages} {41} (\bibinfo {year} {2018})}\BibitemShut
  {NoStop}%
\bibitem [{\citenamefont {Levine}\ \emph {et~al.}(2019)\citenamefont {Levine},
  \citenamefont {Sharir}, \citenamefont {Cohen},\ and\ \citenamefont
  {Shashua}}]{levine_quantum_2019}%
  \BibitemOpen
  \bibfield  {author} {\bibinfo {author} {\bibfnamefont {Y.}~\bibnamefont
  {Levine}}, \bibinfo {author} {\bibfnamefont {O.}~\bibnamefont {Sharir}},
  \bibinfo {author} {\bibfnamefont {N.}~\bibnamefont {Cohen}}, \ and\ \bibinfo
  {author} {\bibfnamefont {A.}~\bibnamefont {Shashua}},\ }\href {\doibase
  10.1103/PhysRevLett.122.065301} {\bibfield  {journal} {\bibinfo  {journal}
  {Physical Review Letters}\ }\textbf {\bibinfo {volume} {122}},\ \bibinfo
  {pages} {065301} (\bibinfo {year} {2019})}\BibitemShut {NoStop}%
\bibitem [{\citenamefont {Carleo}\ \emph {et~al.}(2019)\citenamefont {Carleo},
  \citenamefont {Cirac}, \citenamefont {Cranmer}, \citenamefont {Daudet},
  \citenamefont {Schuld}, \citenamefont {Tishby}, \citenamefont
  {Vogt-Maranto},\ and\ \citenamefont {Zdeborová}}]{carleo_machine_2019}%
  \BibitemOpen
  \bibfield  {author} {\bibinfo {author} {\bibfnamefont {G.}~\bibnamefont
  {Carleo}}, \bibinfo {author} {\bibfnamefont {I.}~\bibnamefont {Cirac}},
  \bibinfo {author} {\bibfnamefont {K.}~\bibnamefont {Cranmer}}, \bibinfo
  {author} {\bibfnamefont {L.}~\bibnamefont {Daudet}}, \bibinfo {author}
  {\bibfnamefont {M.}~\bibnamefont {Schuld}}, \bibinfo {author} {\bibfnamefont
  {N.}~\bibnamefont {Tishby}}, \bibinfo {author} {\bibfnamefont
  {L.}~\bibnamefont {Vogt-Maranto}}, \ and\ \bibinfo {author} {\bibfnamefont
  {L.}~\bibnamefont {Zdeborová}},\ }\href {\doibase
  10.1103/RevModPhys.91.045002} {\bibfield  {journal} {\bibinfo  {journal}
  {Reviews of Modern Physics}\ }\textbf {\bibinfo {volume} {91}},\ \bibinfo
  {pages} {045002} (\bibinfo {year} {2019})}\BibitemShut {NoStop}%
\bibitem [{\citenamefont {Mehta}\ \emph {et~al.}(2019)\citenamefont {Mehta},
  \citenamefont {Bukov}, \citenamefont {Wang}, \citenamefont {Day},
  \citenamefont {Richardson}, \citenamefont {Fisher},\ and\ \citenamefont
  {Schwab}}]{mehta_high-bias_2019}%
  \BibitemOpen
  \bibfield  {author} {\bibinfo {author} {\bibfnamefont {P.}~\bibnamefont
  {Mehta}}, \bibinfo {author} {\bibfnamefont {M.}~\bibnamefont {Bukov}},
  \bibinfo {author} {\bibfnamefont {C.-H.}\ \bibnamefont {Wang}}, \bibinfo
  {author} {\bibfnamefont {A.~G.~R.}\ \bibnamefont {Day}}, \bibinfo {author}
  {\bibfnamefont {C.}~\bibnamefont {Richardson}}, \bibinfo {author}
  {\bibfnamefont {C.~K.}\ \bibnamefont {Fisher}}, \ and\ \bibinfo {author}
  {\bibfnamefont {D.~J.}\ \bibnamefont {Schwab}},\ }\href {\doibase
  10.1016/j.physrep.2019.03.001} {\bibfield  {journal} {\bibinfo  {journal}
  {Physics Reports}\ }\textbf {\bibinfo {volume} {810}},\ \bibinfo {pages} {1}
  (\bibinfo {year} {2019})}\BibitemShut {NoStop}%
\bibitem [{\citenamefont {Breiman}(1995)}]{breiman1995reflections}%
  \BibitemOpen
  \bibfield  {author} {\bibinfo {author} {\bibfnamefont {L.}~\bibnamefont
  {Breiman}},\ }\href@noop {} {\bibfield  {journal} {\bibinfo  {journal} {The
  Mathematics of Generalization}\ }\textbf {\bibinfo {volume} {XX}},\ \bibinfo
  {pages} {11} (\bibinfo {year} {1995})}\BibitemShut {NoStop}%
\bibitem [{\citenamefont {Zdeborov{\'a}}(2020)}]{zdeborova2020understanding}%
  \BibitemOpen
  \bibfield  {author} {\bibinfo {author} {\bibfnamefont {L.}~\bibnamefont
  {Zdeborov{\'a}}},\ }\href {\doibase 10.1038/s41567-020-0929-2} {\bibfield
  {journal} {\bibinfo  {journal} {Nature Physics}\ }\textbf {\bibinfo {volume}
  {16}},\ \bibinfo {pages} {1} (\bibinfo {year} {2020})}\BibitemShut {NoStop}%
\bibitem [{\citenamefont {Zhang}\ \emph {et~al.}(2017)\citenamefont {Zhang},
  \citenamefont {Bengio}, \citenamefont {Hardt}, \citenamefont {Recht},\ and\
  \citenamefont {Vinyals}}]{zhang_understanding_2017}%
  \BibitemOpen
  \bibfield  {author} {\bibinfo {author} {\bibfnamefont {C.}~\bibnamefont
  {Zhang}}, \bibinfo {author} {\bibfnamefont {S.}~\bibnamefont {Bengio}},
  \bibinfo {author} {\bibfnamefont {M.}~\bibnamefont {Hardt}}, \bibinfo
  {author} {\bibfnamefont {B.}~\bibnamefont {Recht}}, \ and\ \bibinfo {author}
  {\bibfnamefont {O.}~\bibnamefont {Vinyals}},\ }in\ \href@noop {} {\emph
  {\bibinfo {booktitle} {The International Conference on Learning
  Representations}}}\ (\bibinfo {year} {2017})\BibitemShut {NoStop}%
\bibitem [{\citenamefont {Simonyan}\ and\ \citenamefont
  {Zisserman}(2014)}]{simonyan2014very}%
  \BibitemOpen
  \bibfield  {author} {\bibinfo {author} {\bibfnamefont {K.}~\bibnamefont
  {Simonyan}}\ and\ \bibinfo {author} {\bibfnamefont {A.}~\bibnamefont
  {Zisserman}},\ }\href@noop {} {\bibfield  {journal} {\bibinfo  {journal}
  {arXiv preprint arXiv:1409.1556}\ } (\bibinfo {year} {2014})}\BibitemShut
  {NoStop}%
\bibitem [{\citenamefont {Brown}\ \emph {et~al.}(2020)\citenamefont {Brown},
  \citenamefont {Mann}, \citenamefont {Ryder}, \citenamefont {Subbiah},
  \citenamefont {Kaplan}, \citenamefont {Dhariwal}, \citenamefont
  {Neelakantan}, \citenamefont {Shyam}, \citenamefont {Sastry}, \citenamefont
  {Askell} \emph {et~al.}}]{brown2020language}%
  \BibitemOpen
  \bibfield  {author} {\bibinfo {author} {\bibfnamefont {T.~B.}\ \bibnamefont
  {Brown}}, \bibinfo {author} {\bibfnamefont {B.}~\bibnamefont {Mann}},
  \bibinfo {author} {\bibfnamefont {N.}~\bibnamefont {Ryder}}, \bibinfo
  {author} {\bibfnamefont {M.}~\bibnamefont {Subbiah}}, \bibinfo {author}
  {\bibfnamefont {J.}~\bibnamefont {Kaplan}}, \bibinfo {author} {\bibfnamefont
  {P.}~\bibnamefont {Dhariwal}}, \bibinfo {author} {\bibfnamefont
  {A.}~\bibnamefont {Neelakantan}}, \bibinfo {author} {\bibfnamefont
  {P.}~\bibnamefont {Shyam}}, \bibinfo {author} {\bibfnamefont
  {G.}~\bibnamefont {Sastry}}, \bibinfo {author} {\bibfnamefont
  {A.}~\bibnamefont {Askell}},  \emph {et~al.},\ }\href@noop {} {\bibfield
  {journal} {\bibinfo  {journal} {arXiv preprint arXiv:2005.14165}\ } (\bibinfo
  {year} {2020})}\BibitemShut {NoStop}%
\bibitem [{\citenamefont {Dyson}(2004)}]{dyson_meeting_2004}%
  \BibitemOpen
  \bibfield  {author} {\bibinfo {author} {\bibfnamefont {F.}~\bibnamefont
  {Dyson}},\ }\href {\doibase 10.1038/427297a} {\bibfield  {journal} {\bibinfo
  {journal} {Nature}\ }\textbf {\bibinfo {volume} {427}},\ \bibinfo {pages}
  {297} (\bibinfo {year} {2004})}\BibitemShut {NoStop}%
\bibitem [{\citenamefont {Saxe}\ \emph {et~al.}(2014)\citenamefont {Saxe},
  \citenamefont {McClelland},\ and\ \citenamefont {Ganguli}}]{saxe_exact_2013}%
  \BibitemOpen
  \bibfield  {author} {\bibinfo {author} {\bibfnamefont {A.~M.}\ \bibnamefont
  {Saxe}}, \bibinfo {author} {\bibfnamefont {J.~L.}\ \bibnamefont
  {McClelland}}, \ and\ \bibinfo {author} {\bibfnamefont {S.}~\bibnamefont
  {Ganguli}},\ }in\ \href@noop {} {\emph {\bibinfo {booktitle} {The
  International Conference on Learning Representations}}}\ (\bibinfo {year}
  {2014})\BibitemShut {NoStop}%
\bibitem [{\citenamefont {Saxe}\ \emph {et~al.}(2019)\citenamefont {Saxe},
  \citenamefont {Bansal}, \citenamefont {Dapello}, \citenamefont {Advani},
  \citenamefont {Kolchinsky}, \citenamefont {Tracey},\ and\ \citenamefont
  {Cox}}]{saxe_information_2019}%
  \BibitemOpen
  \bibfield  {author} {\bibinfo {author} {\bibfnamefont {A.~M.}\ \bibnamefont
  {Saxe}}, \bibinfo {author} {\bibfnamefont {Y.}~\bibnamefont {Bansal}},
  \bibinfo {author} {\bibfnamefont {J.}~\bibnamefont {Dapello}}, \bibinfo
  {author} {\bibfnamefont {M.}~\bibnamefont {Advani}}, \bibinfo {author}
  {\bibfnamefont {A.}~\bibnamefont {Kolchinsky}}, \bibinfo {author}
  {\bibfnamefont {B.~D.}\ \bibnamefont {Tracey}}, \ and\ \bibinfo {author}
  {\bibfnamefont {D.~D.}\ \bibnamefont {Cox}},\ }\href {\doibase
  10.1088/1742-5468/ab3985} {\bibfield  {journal} {\bibinfo  {journal} {Journal
  of Statistical Mechanics: Theory and Experiment}\ }\textbf {\bibinfo {volume}
  {2019}},\ \bibinfo {pages} {124020} (\bibinfo {year} {2019})}\BibitemShut
  {NoStop}%
\bibitem [{\citenamefont {Lampinen}\ and\ \citenamefont
  {Ganguli}(2019)}]{lampinen_analytic_2019}%
  \BibitemOpen
  \bibfield  {author} {\bibinfo {author} {\bibfnamefont {A.~K.}\ \bibnamefont
  {Lampinen}}\ and\ \bibinfo {author} {\bibfnamefont {S.}~\bibnamefont
  {Ganguli}},\ }in\ \href@noop {} {\emph {\bibinfo {booktitle} {The
  International Conference on Learning Representations}}}\ (\bibinfo {year}
  {2019})\BibitemShut {NoStop}%
\bibitem [{\citenamefont {Engel}\ and\ \citenamefont
  {Broeck}(2001)}]{engel_statistical_2001}%
  \BibitemOpen
  \bibfield  {author} {\bibinfo {author} {\bibfnamefont {A.}~\bibnamefont
  {Engel}}\ and\ \bibinfo {author} {\bibfnamefont {C.~V.~d.}\ \bibnamefont
  {Broeck}},\ }\href@noop {} {\emph {\bibinfo {title} {Statistical {Mechanics}
  of {Learning}}}}\ (\bibinfo {year} {2001})\ \bibinfo {note} {google-Books-ID:
  qVo4IT9ByfQC}\BibitemShut {NoStop}%
\bibitem [{\citenamefont {Aubin}\ \emph {et~al.}(2018)\citenamefont {Aubin},
  \citenamefont {Maillard}, \citenamefont {barbier}, \citenamefont {Krzakala},
  \citenamefont {Macris},\ and\ \citenamefont
  {Zdeborová}}]{aubin_committee_2018}%
  \BibitemOpen
  \bibfield  {author} {\bibinfo {author} {\bibfnamefont {B.}~\bibnamefont
  {Aubin}}, \bibinfo {author} {\bibfnamefont {A.}~\bibnamefont {Maillard}},
  \bibinfo {author} {\bibfnamefont {j.}~\bibnamefont {barbier}}, \bibinfo
  {author} {\bibfnamefont {F.}~\bibnamefont {Krzakala}}, \bibinfo {author}
  {\bibfnamefont {N.}~\bibnamefont {Macris}}, \ and\ \bibinfo {author}
  {\bibfnamefont {L.}~\bibnamefont {Zdeborová}},\ }in\ \href@noop {} {\emph
  {\bibinfo {booktitle} {Advances in {Neural} {Information} {Processing}
  {Systems} 31}}}\ (\bibinfo {year} {2018})\ pp.\ \bibinfo {pages}
  {3223--3234}\BibitemShut {NoStop}%
\bibitem [{\citenamefont {Choromanska}\ \emph {et~al.}(2015)\citenamefont
  {Choromanska}, \citenamefont {Henaff}, \citenamefont {Mathieu}, \citenamefont
  {Arous},\ and\ \citenamefont {LeCun}}]{choromanska_loss_2015}%
  \BibitemOpen
  \bibfield  {author} {\bibinfo {author} {\bibfnamefont {A.}~\bibnamefont
  {Choromanska}}, \bibinfo {author} {\bibfnamefont {M.}~\bibnamefont {Henaff}},
  \bibinfo {author} {\bibfnamefont {M.}~\bibnamefont {Mathieu}}, \bibinfo
  {author} {\bibfnamefont {G.~B.}\ \bibnamefont {Arous}}, \ and\ \bibinfo
  {author} {\bibfnamefont {Y.}~\bibnamefont {LeCun}},\ }in\ \href@noop {}
  {\emph {\bibinfo {booktitle} {Artificial Intelligence and Statistics}}}\
  (\bibinfo {year} {2015})\ pp.\ \bibinfo {pages} {192--204}\BibitemShut
  {NoStop}%
\bibitem [{\citenamefont {Mei}\ \emph {et~al.}(2018)\citenamefont {Mei},
  \citenamefont {Montanari},\ and\ \citenamefont {Nguyen}}]{mei_mean_2018}%
  \BibitemOpen
  \bibfield  {author} {\bibinfo {author} {\bibfnamefont {S.}~\bibnamefont
  {Mei}}, \bibinfo {author} {\bibfnamefont {A.}~\bibnamefont {Montanari}}, \
  and\ \bibinfo {author} {\bibfnamefont {P.-M.}\ \bibnamefont {Nguyen}},\
  }\href {\doibase 10.1073/pnas.1806579115} {\bibfield  {journal} {\bibinfo
  {journal} {Proceedings of the National Academy of Sciences}\ }\textbf
  {\bibinfo {volume} {115}},\ \bibinfo {pages} {E7665} (\bibinfo {year}
  {2018})}\BibitemShut {NoStop}%
\bibitem [{\citenamefont {Rotskoff}\ and\ \citenamefont
  {Vanden-Eijnden}(2018)}]{rotskoff_parameters_2018}%
  \BibitemOpen
  \bibfield  {author} {\bibinfo {author} {\bibfnamefont {G.}~\bibnamefont
  {Rotskoff}}\ and\ \bibinfo {author} {\bibfnamefont {E.}~\bibnamefont
  {Vanden-Eijnden}},\ }in\ \href@noop {} {\emph {\bibinfo {booktitle} {Advances
  in {Neural} {Information} {Processing} {Systems} 31}}}\ (\bibinfo {year}
  {2018})\ pp.\ \bibinfo {pages} {7146--7155}\BibitemShut {NoStop}%
\bibitem [{\citenamefont {Chizat}\ and\ \citenamefont
  {Bach}(2018)}]{chizat_global_2018}%
  \BibitemOpen
  \bibfield  {author} {\bibinfo {author} {\bibfnamefont {L.}~\bibnamefont
  {Chizat}}\ and\ \bibinfo {author} {\bibfnamefont {F.}~\bibnamefont {Bach}},\
  }in\ \href@noop {} {\emph {\bibinfo {booktitle} {Advances in {Neural}
  {Information} {Processing} {Systems} 31}}}\ (\bibinfo {year} {2018})\ pp.\
  \bibinfo {pages} {3036--3046}\BibitemShut {NoStop}%
\bibitem [{\citenamefont {Sirignano}\ and\ \citenamefont
  {Spiliopoulos}(2020)}]{sirignano_mean_2020}%
  \BibitemOpen
  \bibfield  {author} {\bibinfo {author} {\bibfnamefont {J.}~\bibnamefont
  {Sirignano}}\ and\ \bibinfo {author} {\bibfnamefont {K.}~\bibnamefont
  {Spiliopoulos}},\ }\href {\doibase 10.1016/j.spa.2019.06.003} {\bibfield
  {journal} {\bibinfo  {journal} {Stochastic Processes and their Applications}\
  }\textbf {\bibinfo {volume} {130}},\ \bibinfo {pages} {1820} (\bibinfo {year}
  {2020})}\BibitemShut {NoStop}%
\bibitem [{\citenamefont {Jacot}\ \emph {et~al.}(2018)\citenamefont {Jacot},
  \citenamefont {Gabriel},\ and\ \citenamefont {Hongler}}]{jacot_neural_2018}%
  \BibitemOpen
  \bibfield  {author} {\bibinfo {author} {\bibfnamefont {A.}~\bibnamefont
  {Jacot}}, \bibinfo {author} {\bibfnamefont {F.}~\bibnamefont {Gabriel}}, \
  and\ \bibinfo {author} {\bibfnamefont {C.}~\bibnamefont {Hongler}},\ }in\
  \href@noop {} {\emph {\bibinfo {booktitle} {Advances in {Neural}
  {Information} {Processing} {Systems} 31}}}\ (\bibinfo {year} {2018})\ pp.\
  \bibinfo {pages} {8571--8580}\BibitemShut {NoStop}%
\bibitem [{\citenamefont {Lee}\ \emph {et~al.}(2019)\citenamefont {Lee},
  \citenamefont {Xiao}, \citenamefont {Schoenholz}, \citenamefont {Bahri},
  \citenamefont {Novak}, \citenamefont {Sohl-Dickstein},\ and\ \citenamefont
  {Pennington}}]{lee_wide_2019}%
  \BibitemOpen
  \bibfield  {author} {\bibinfo {author} {\bibfnamefont {J.}~\bibnamefont
  {Lee}}, \bibinfo {author} {\bibfnamefont {L.}~\bibnamefont {Xiao}}, \bibinfo
  {author} {\bibfnamefont {S.}~\bibnamefont {Schoenholz}}, \bibinfo {author}
  {\bibfnamefont {Y.}~\bibnamefont {Bahri}}, \bibinfo {author} {\bibfnamefont
  {R.}~\bibnamefont {Novak}}, \bibinfo {author} {\bibfnamefont
  {J.}~\bibnamefont {Sohl-Dickstein}}, \ and\ \bibinfo {author} {\bibfnamefont
  {J.}~\bibnamefont {Pennington}},\ }in\ \href@noop {} {\emph {\bibinfo
  {booktitle} {Advances in {Neural} {Information} {Processing} {Systems} 32}}}\
  (\bibinfo {year} {2019})\ pp.\ \bibinfo {pages} {8572--8583}\BibitemShut
  {NoStop}%
\bibitem [{\citenamefont {Arpit}\ \emph {et~al.}(2017)\citenamefont {Arpit},
  \citenamefont {Jastrzbski}, \citenamefont {Ballas}, \citenamefont {Krueger},
  \citenamefont {Bengio}, \citenamefont {Kanwal}, \citenamefont {Maharaj},
  \citenamefont {Fischer}, \citenamefont {Courville}, \citenamefont {Bengio}
  \emph {et~al.}}]{arpit2017closer}%
  \BibitemOpen
  \bibfield  {author} {\bibinfo {author} {\bibfnamefont {D.}~\bibnamefont
  {Arpit}}, \bibinfo {author} {\bibfnamefont {S.}~\bibnamefont {Jastrzbski}},
  \bibinfo {author} {\bibfnamefont {N.}~\bibnamefont {Ballas}}, \bibinfo
  {author} {\bibfnamefont {D.}~\bibnamefont {Krueger}}, \bibinfo {author}
  {\bibfnamefont {E.}~\bibnamefont {Bengio}}, \bibinfo {author} {\bibfnamefont
  {M.~S.}\ \bibnamefont {Kanwal}}, \bibinfo {author} {\bibfnamefont
  {T.}~\bibnamefont {Maharaj}}, \bibinfo {author} {\bibfnamefont
  {A.}~\bibnamefont {Fischer}}, \bibinfo {author} {\bibfnamefont
  {A.}~\bibnamefont {Courville}}, \bibinfo {author} {\bibfnamefont
  {Y.}~\bibnamefont {Bengio}},  \emph {et~al.},\ }in\ \href@noop {} {\emph
  {\bibinfo {booktitle} {Proceedings of the 34th International Conference on
  Machine Learning-Volume 70}}}\ (\bibinfo {year} {2017})\ pp.\ \bibinfo
  {pages} {233--242}\BibitemShut {NoStop}%
\bibitem [{\citenamefont {Kalimeris}\ \emph {et~al.}(2019)\citenamefont
  {Kalimeris}, \citenamefont {Kaplun}, \citenamefont {Nakkiran}, \citenamefont
  {Edelman}, \citenamefont {Yang}, \citenamefont {Barak},\ and\ \citenamefont
  {Zhang}}]{kalimeris_sgd_2019}%
  \BibitemOpen
  \bibfield  {author} {\bibinfo {author} {\bibfnamefont {D.}~\bibnamefont
  {Kalimeris}}, \bibinfo {author} {\bibfnamefont {G.}~\bibnamefont {Kaplun}},
  \bibinfo {author} {\bibfnamefont {P.}~\bibnamefont {Nakkiran}}, \bibinfo
  {author} {\bibfnamefont {B.}~\bibnamefont {Edelman}}, \bibinfo {author}
  {\bibfnamefont {T.}~\bibnamefont {Yang}}, \bibinfo {author} {\bibfnamefont
  {B.}~\bibnamefont {Barak}}, \ and\ \bibinfo {author} {\bibfnamefont
  {H.}~\bibnamefont {Zhang}},\ }in\ \href@noop {} {\emph {\bibinfo {booktitle}
  {Advances in {Neural} {Information} {Processing} {Systems} 32}}}\ (\bibinfo
  {year} {2019})\ pp.\ \bibinfo {pages} {3496--3506}\BibitemShut {NoStop}%
\bibitem [{\citenamefont {Valle-Perez}\ \emph {et~al.}(2019)\citenamefont
  {Valle-Perez}, \citenamefont {Camargo},\ and\ \citenamefont
  {Louis}}]{valle2018deep}%
  \BibitemOpen
  \bibfield  {author} {\bibinfo {author} {\bibfnamefont {G.}~\bibnamefont
  {Valle-Perez}}, \bibinfo {author} {\bibfnamefont {C.~Q.}\ \bibnamefont
  {Camargo}}, \ and\ \bibinfo {author} {\bibfnamefont {A.~A.}\ \bibnamefont
  {Louis}},\ }in\ \href@noop {} {\emph {\bibinfo {booktitle} {The International
  Conference on Learning Representations}}}\ (\bibinfo {year}
  {2019})\BibitemShut {NoStop}%
\bibitem [{\citenamefont {Xu}\ \emph {et~al.}(2019)\citenamefont {Xu},
  \citenamefont {Zhang},\ and\ \citenamefont {Xiao}}]{xu_training_2019}%
  \BibitemOpen
  \bibfield  {author} {\bibinfo {author} {\bibfnamefont {Z.-Q.~J.}\
  \bibnamefont {Xu}}, \bibinfo {author} {\bibfnamefont {Y.}~\bibnamefont
  {Zhang}}, \ and\ \bibinfo {author} {\bibfnamefont {Y.}~\bibnamefont {Xiao}},\
  }in\ \href {\doibase 10.1007/978-3-030-36708-4_22} {\emph {\bibinfo
  {booktitle} {Neural {Information} {Processing}}}},\ \bibinfo {series and
  number} {Lecture {Notes} in {Computer} {Science}}\ (\bibinfo {year} {2019})\
  pp.\ \bibinfo {pages} {264--274}\BibitemShut {NoStop}%
\bibitem [{\citenamefont {Xu}\ \emph {et~al.}(2020)\citenamefont {Xu},
  \citenamefont {Zhang}, \citenamefont {Luo}, \citenamefont {Xiao},\ and\
  \citenamefont {Ma}}]{xu_frequency_2019}%
  \BibitemOpen
  \bibfield  {author} {\bibinfo {author} {\bibfnamefont {Z.-Q.~J.}\
  \bibnamefont {Xu}}, \bibinfo {author} {\bibfnamefont {Y.}~\bibnamefont
  {Zhang}}, \bibinfo {author} {\bibfnamefont {T.}~\bibnamefont {Luo}}, \bibinfo
  {author} {\bibfnamefont {Y.}~\bibnamefont {Xiao}}, \ and\ \bibinfo {author}
  {\bibfnamefont {Z.}~\bibnamefont {Ma}},\ }\href@noop {} {\bibfield  {journal}
  {\bibinfo  {journal} {Communications in Computational Physics}\ }\textbf
  {\bibinfo {volume} {28}},\ \bibinfo {pages} {1746} (\bibinfo {year}
  {2020})}\BibitemShut {NoStop}%
\bibitem [{\citenamefont {Rahaman}\ \emph {et~al.}(2019)\citenamefont
  {Rahaman}, \citenamefont {Baratin}, \citenamefont {Arpit}, \citenamefont
  {Draxler}, \citenamefont {Lin}, \citenamefont {Hamprecht}, \citenamefont
  {Bengio},\ and\ \citenamefont {Courville}}]{rahaman_spectral_2019}%
  \BibitemOpen
  \bibfield  {author} {\bibinfo {author} {\bibfnamefont {N.}~\bibnamefont
  {Rahaman}}, \bibinfo {author} {\bibfnamefont {A.}~\bibnamefont {Baratin}},
  \bibinfo {author} {\bibfnamefont {D.}~\bibnamefont {Arpit}}, \bibinfo
  {author} {\bibfnamefont {F.}~\bibnamefont {Draxler}}, \bibinfo {author}
  {\bibfnamefont {M.}~\bibnamefont {Lin}}, \bibinfo {author} {\bibfnamefont
  {F.}~\bibnamefont {Hamprecht}}, \bibinfo {author} {\bibfnamefont
  {Y.}~\bibnamefont {Bengio}}, \ and\ \bibinfo {author} {\bibfnamefont
  {A.}~\bibnamefont {Courville}},\ }in\ \href@noop {} {\emph {\bibinfo
  {booktitle} {International {Conference} on {Machine} {Learning}}}}\ (\bibinfo
  {year} {2019})\ pp.\ \bibinfo {pages} {5301--5310}\BibitemShut {NoStop}%
\bibitem [{\citenamefont {Ronen}\ \emph
  {et~al.}(2019{\natexlab{a}})\citenamefont {Ronen}, \citenamefont {Jacobs},
  \citenamefont {Kasten},\ and\ \citenamefont
  {Kritchman}}]{basri2019convergence}%
  \BibitemOpen
  \bibfield  {author} {\bibinfo {author} {\bibfnamefont {B.}~\bibnamefont
  {Ronen}}, \bibinfo {author} {\bibfnamefont {D.}~\bibnamefont {Jacobs}},
  \bibinfo {author} {\bibfnamefont {Y.}~\bibnamefont {Kasten}}, \ and\ \bibinfo
  {author} {\bibfnamefont {S.}~\bibnamefont {Kritchman}},\ }in\ \href@noop {}
  {\emph {\bibinfo {booktitle} {Advances in Neural Information Processing
  Systems}}}\ (\bibinfo {year} {2019})\ pp.\ \bibinfo {pages}
  {4763--4772}\BibitemShut {NoStop}%
\bibitem [{\citenamefont {Rabinowitz}(2019)}]{rabinowitz_meta-learners_2019}%
  \BibitemOpen
  \bibfield  {author} {\bibinfo {author} {\bibfnamefont {N.~C.}\ \bibnamefont
  {Rabinowitz}},\ }\href@noop {} {\bibfield  {journal} {\bibinfo  {journal}
  {arXiv:1905.01320 [cs, stat]}\ } (\bibinfo {year} {2019})}\BibitemShut
  {NoStop}%
\bibitem [{\citenamefont {Jagtap}\ \emph {et~al.}(2020)\citenamefont {Jagtap},
  \citenamefont {Kawaguchi},\ and\ \citenamefont
  {Karniadakis}}]{jagtap_adaptive_2020}%
  \BibitemOpen
  \bibfield  {author} {\bibinfo {author} {\bibfnamefont {A.~D.}\ \bibnamefont
  {Jagtap}}, \bibinfo {author} {\bibfnamefont {K.}~\bibnamefont {Kawaguchi}}, \
  and\ \bibinfo {author} {\bibfnamefont {G.~E.}\ \bibnamefont {Karniadakis}},\
  }\href {\doibase 10.1016/j.jcp.2019.109136} {\bibfield  {journal} {\bibinfo
  {journal} {Journal of Computational Physics}\ }\textbf {\bibinfo {volume}
  {404}},\ \bibinfo {pages} {109136} (\bibinfo {year} {2020})}\BibitemShut
  {NoStop}%
\bibitem [{\citenamefont {Ronen}\ \emph
  {et~al.}(2019{\natexlab{b}})\citenamefont {Ronen}, \citenamefont {Jacobs},
  \citenamefont {Kasten},\ and\ \citenamefont
  {Kritchman}}]{ronen_convergence_2019}%
  \BibitemOpen
  \bibfield  {author} {\bibinfo {author} {\bibfnamefont {B.}~\bibnamefont
  {Ronen}}, \bibinfo {author} {\bibfnamefont {D.}~\bibnamefont {Jacobs}},
  \bibinfo {author} {\bibfnamefont {Y.}~\bibnamefont {Kasten}}, \ and\ \bibinfo
  {author} {\bibfnamefont {S.}~\bibnamefont {Kritchman}},\ }in\ \href@noop {}
  {\emph {\bibinfo {booktitle} {Advances in {Neural} {Information} {Processing}
  {Systems} 32}}}\ (\bibinfo {year} {2019})\ pp.\ \bibinfo {pages}
  {4761--4771}\BibitemShut {NoStop}%
\bibitem [{\citenamefont {Yang}\ and\ \citenamefont
  {Salman}(2020)}]{yang_fine-grained_2020}%
  \BibitemOpen
  \bibfield  {author} {\bibinfo {author} {\bibfnamefont {G.}~\bibnamefont
  {Yang}}\ and\ \bibinfo {author} {\bibfnamefont {H.}~\bibnamefont {Salman}},\
  }\href@noop {} {\bibfield  {journal} {\bibinfo  {journal} {arXiv:1907.10599
  [cs, stat]}\ } (\bibinfo {year} {2020})}\BibitemShut {NoStop}%
\bibitem [{\citenamefont {Cao}\ \emph {et~al.}(2019)\citenamefont {Cao},
  \citenamefont {Fang}, \citenamefont {Wu}, \citenamefont {Zhou},\ and\
  \citenamefont {Gu}}]{cao2019towards}%
  \BibitemOpen
  \bibfield  {author} {\bibinfo {author} {\bibfnamefont {Y.}~\bibnamefont
  {Cao}}, \bibinfo {author} {\bibfnamefont {Z.}~\bibnamefont {Fang}}, \bibinfo
  {author} {\bibfnamefont {Y.}~\bibnamefont {Wu}}, \bibinfo {author}
  {\bibfnamefont {D.-X.}\ \bibnamefont {Zhou}}, \ and\ \bibinfo {author}
  {\bibfnamefont {Q.}~\bibnamefont {Gu}},\ }\href@noop {} {\bibfield  {journal}
  {\bibinfo  {journal} {arXiv preprint arXiv:1912.01198}\ } (\bibinfo {year}
  {2019})}\BibitemShut {NoStop}%
\bibitem [{\citenamefont {Cai}\ \emph {et~al.}(2019)\citenamefont {Cai},
  \citenamefont {Li},\ and\ \citenamefont {Liu}}]{cai_phase_2019}%
  \BibitemOpen
  \bibfield  {author} {\bibinfo {author} {\bibfnamefont {W.}~\bibnamefont
  {Cai}}, \bibinfo {author} {\bibfnamefont {X.}~\bibnamefont {Li}}, \ and\
  \bibinfo {author} {\bibfnamefont {L.}~\bibnamefont {Liu}},\ }\href@noop {}
  {\bibfield  {journal} {\bibinfo  {journal} {arXiv:1909.11759 To appear in
  SIAM J. Scientific Computing}\ } (\bibinfo {year} {2019})}\BibitemShut
  {NoStop}%
\bibitem [{\citenamefont {Biland}\ \emph {et~al.}(2019)\citenamefont {Biland},
  \citenamefont {Azevedo}, \citenamefont {Kim},\ and\ \citenamefont
  {Solenthaler}}]{biland_frequency-aware_2019}%
  \BibitemOpen
  \bibfield  {author} {\bibinfo {author} {\bibfnamefont {S.}~\bibnamefont
  {Biland}}, \bibinfo {author} {\bibfnamefont {V.~C.}\ \bibnamefont {Azevedo}},
  \bibinfo {author} {\bibfnamefont {B.}~\bibnamefont {Kim}}, \ and\ \bibinfo
  {author} {\bibfnamefont {B.}~\bibnamefont {Solenthaler}},\ }\href@noop {}
  {\bibfield  {journal} {\bibinfo  {journal} {arXiv:1912.08776 [physics,
  stat]}\ } (\bibinfo {year} {2019})}\BibitemShut {NoStop}%
\bibitem [{\citenamefont {Liu}\ \emph {et~al.}(2020)\citenamefont {Liu},
  \citenamefont {Cai},\ and\ \citenamefont {Xu}}]{liu2019multi}%
  \BibitemOpen
  \bibfield  {author} {\bibinfo {author} {\bibfnamefont {Z.}~\bibnamefont
  {Liu}}, \bibinfo {author} {\bibfnamefont {W.}~\bibnamefont {Cai}}, \ and\
  \bibinfo {author} {\bibfnamefont {Z.-Q.~J.}\ \bibnamefont {Xu}},\ }\href@noop
  {} {\bibfield  {journal} {\bibinfo  {journal} {Communications in
  Computational Physics}\ }\textbf {\bibinfo {volume} {28}},\ \bibinfo {pages}
  {1970} (\bibinfo {year} {2020})}\BibitemShut {NoStop}%
\bibitem [{\citenamefont {Li}\ \emph {et~al.}(2020)\citenamefont {Li},
  \citenamefont {Xu},\ and\ \citenamefont {Zhang}}]{li2019multi}%
  \BibitemOpen
  \bibfield  {author} {\bibinfo {author} {\bibfnamefont {X.-A.}\ \bibnamefont
  {Li}}, \bibinfo {author} {\bibfnamefont {Z.-Q.~J.}\ \bibnamefont {Xu}}, \
  and\ \bibinfo {author} {\bibfnamefont {L.}~\bibnamefont {Zhang}},\
  }\href@noop {} {\bibfield  {journal} {\bibinfo  {journal} {Communications in
  Computational Physics}\ }\textbf {\bibinfo {volume} {28}},\ \bibinfo {pages}
  {1886} (\bibinfo {year} {2020})}\BibitemShut {NoStop}%
\bibitem [{\citenamefont {Wang}\ \emph {et~al.}(2020)\citenamefont {Wang},
  \citenamefont {Zhang},\ and\ \citenamefont {Cai}}]{wang2019multi}%
  \BibitemOpen
  \bibfield  {author} {\bibinfo {author} {\bibfnamefont {B.}~\bibnamefont
  {Wang}}, \bibinfo {author} {\bibfnamefont {W.}~\bibnamefont {Zhang}}, \ and\
  \bibinfo {author} {\bibfnamefont {W.}~\bibnamefont {Cai}},\ }\href@noop {}
  {\bibfield  {journal} {\bibinfo  {journal} {Communications in Computational
  Physics}\ }\textbf {\bibinfo {volume} {28}},\ \bibinfo {pages} {2139}
  (\bibinfo {year} {2020})}\BibitemShut {NoStop}%
\bibitem [{\citenamefont {Zhang}\ \emph {et~al.}(2019)\citenamefont {Zhang},
  \citenamefont {Xu}, \citenamefont {Luo},\ and\ \citenamefont
  {Ma}}]{zhang_type_2019}%
  \BibitemOpen
  \bibfield  {author} {\bibinfo {author} {\bibfnamefont {Y.}~\bibnamefont
  {Zhang}}, \bibinfo {author} {\bibfnamefont {Z.-Q.~J.}\ \bibnamefont {Xu}},
  \bibinfo {author} {\bibfnamefont {T.}~\bibnamefont {Luo}}, \ and\ \bibinfo
  {author} {\bibfnamefont {Z.}~\bibnamefont {Ma}},\ }\href@noop {} {\bibfield
  {journal} {\bibinfo  {journal} {arXiv:1905.07777 [cs, stat]}\ } (\bibinfo
  {year} {2019})}\BibitemShut {NoStop}%
\bibitem [{\citenamefont {E}\ \emph {et~al.}(2019)\citenamefont {E},
  \citenamefont {Ma},\ and\ \citenamefont {Wu}}]{e_priori_2019}%
  \BibitemOpen
  \bibfield  {author} {\bibinfo {author} {\bibfnamefont {W.}~\bibnamefont {E}},
  \bibinfo {author} {\bibfnamefont {C.}~\bibnamefont {Ma}}, \ and\ \bibinfo
  {author} {\bibfnamefont {L.}~\bibnamefont {Wu}},\ }\href {\doibase
  10.4310/CMS.2019.v17.n5.a11} {\bibfield  {journal} {\bibinfo  {journal}
  {Communications in Mathematical Sciences}\ }\textbf {\bibinfo {volume}
  {17}},\ \bibinfo {pages} {1407} (\bibinfo {year} {2019})}\BibitemShut
  {NoStop}%
\bibitem [{\citenamefont {Minsky}\ and\ \citenamefont
  {Papert}(2017)}]{minsky2017perceptrons}%
  \BibitemOpen
  \bibfield  {author} {\bibinfo {author} {\bibfnamefont {M.}~\bibnamefont
  {Minsky}}\ and\ \bibinfo {author} {\bibfnamefont {S.~A.}\ \bibnamefont
  {Papert}},\ }\href@noop {} {\emph {\bibinfo {title} {Perceptrons: An
  introduction to computational geometry}}}\ (\bibinfo  {publisher} {MIT
  press},\ \bibinfo {year} {2017})\BibitemShut {NoStop}%
\bibitem [{\citenamefont {Allender}(1996)}]{allender1996circuit}%
  \BibitemOpen
  \bibfield  {author} {\bibinfo {author} {\bibfnamefont {E.}~\bibnamefont
  {Allender}},\ }in\ \href@noop {} {\emph {\bibinfo {booktitle} {International
  Conference on Foundations of Software Technology and Theoretical Computer
  Science}}}\ (\bibinfo {organization} {Springer},\ \bibinfo {year} {1996})\
  pp.\ \bibinfo {pages} {1--18}\BibitemShut {NoStop}%
\bibitem [{\citenamefont {Arora}\ \emph {et~al.}(2019)\citenamefont {Arora},
  \citenamefont {Du}, \citenamefont {Hu}, \citenamefont {Li},\ and\
  \citenamefont {Wang}}]{arora2019fine}%
  \BibitemOpen
  \bibfield  {author} {\bibinfo {author} {\bibfnamefont {S.}~\bibnamefont
  {Arora}}, \bibinfo {author} {\bibfnamefont {S.}~\bibnamefont {Du}}, \bibinfo
  {author} {\bibfnamefont {W.}~\bibnamefont {Hu}}, \bibinfo {author}
  {\bibfnamefont {Z.}~\bibnamefont {Li}}, \ and\ \bibinfo {author}
  {\bibfnamefont {R.}~\bibnamefont {Wang}},\ }in\ \href@noop {} {\emph
  {\bibinfo {booktitle} {International Conference on Machine Learning}}}\
  (\bibinfo {year} {2019})\ pp.\ \bibinfo {pages} {322--332}\BibitemShut
  {NoStop}%
\bibitem [{\citenamefont {E}\ \emph {et~al.}(2020)\citenamefont {E},
  \citenamefont {Ma},\ and\ \citenamefont {Wu}}]{e2020comparative}%
  \BibitemOpen
  \bibfield  {author} {\bibinfo {author} {\bibfnamefont {W.}~\bibnamefont {E}},
  \bibinfo {author} {\bibfnamefont {C.}~\bibnamefont {Ma}}, \ and\ \bibinfo
  {author} {\bibfnamefont {L.}~\bibnamefont {Wu}},\ }\href@noop {} {\bibfield
  {journal} {\bibinfo  {journal} {Sci. China Math.}\ }\textbf {\bibinfo
  {volume} {63}} (\bibinfo {year} {2020})}\BibitemShut {NoStop}%
\bibitem [{\citenamefont {Cai}\ and\ \citenamefont
  {Liu}(2018)}]{cai_approximating_2018}%
  \BibitemOpen
  \bibfield  {author} {\bibinfo {author} {\bibfnamefont {Z.}~\bibnamefont
  {Cai}}\ and\ \bibinfo {author} {\bibfnamefont {J.}~\bibnamefont {Liu}},\
  }\href {\doibase 10.1103/PhysRevB.97.035116} {\bibfield  {journal} {\bibinfo
  {journal} {Physical Review B}\ }\textbf {\bibinfo {volume} {97}},\ \bibinfo
  {pages} {035116} (\bibinfo {year} {2018})}\BibitemShut {NoStop}%
\end{thebibliography}%

\end{document}